\documentclass[letterpaper]{article} 
\usepackage{aaai2026}  
\usepackage{times}  
\usepackage{helvet}  
\usepackage{courier}  
\usepackage[hyphens]{url}  
\usepackage{graphicx} 
\usepackage{natbib}  
\usepackage{caption} 
\usepackage{algorithm}
\usepackage{algorithmic}
\usepackage{multirow}
\usepackage{amssymb}
\usepackage{subcaption}
\usepackage{newfloat}
\usepackage{listings}
\DeclareCaptionStyle{ruled}{labelfont=normalfont,labelsep=colon,strut=off} 
\floatstyle{ruled}
\newfloat{listing}{tb}{lst}{}
\floatname{listing}{Listing}
\title{Image-Text Knowledge Modeling for Unsupervised Multi-Scenario Person Re-Identification\footnote{This is a preprint of a paper accepted for AAAI 2026 that is expanded to include Supplementary Material. Copyright will transfer to AAAI for the published paper \cite{Pang:ITKMMultScenReID:AAAI2026}, which should be cited for referencing the work presented here.}}
\author {
    Zhiqi Pang\textsuperscript{\rm 1},
    Lingling Zhao\textsuperscript{\rm 1},
    Yang Liu\textsuperscript{\rm 1},
    Chunyu Wang\textsuperscript{\rm 1}\thanks{Corresponding author: Chunyu Wang},
    Gaurav Sharma\textsuperscript{\rm 2}
}
\affiliations {
    \textsuperscript{\rm 1}Faculty of Computing, Harbin Institute of Technology, Harbin, China \\
    \textsuperscript{\rm 2}Department of Electrical and Computer Engineering, University of Rochester, Rochester, USA\\
    zqpang98@gmail.com, zhaoll@hit.edu.cn, liuyang@hit.edu.cn, chunyu@hit.edu.cn, g.sharma@ieee.org
}

\usepackage{bibentry}

\begin{document}

\maketitle

\begin{abstract}
  We propose unsupervised multi-scenario (UMS) person re-identification (ReID) as a new task that expands ReID across diverse scenarios (cross-resolution, clothing change, etc.) within a single coherent framework. To tackle UMS-ReID, we introduce image-text knowledge modeling (ITKM) -- a three-stage framework that effectively exploits the representational power of vision-language models. We start with a pre-trained CLIP model with an image encoder and a text encoder. In Stage I, we introduce a scenario embedding in the image encoder and fine-tune the encoder to adaptively leverage knowledge from multiple scenarios. In Stage II, we optimize a set of learned text embeddings to associate with pseudo-labels from Stage I and introduce a multi-scenario separation loss to increase the divergence between inter-scenario text representations. In Stage III, we first introduce cluster-level and instance-level heterogeneous matching modules to obtain reliable heterogeneous positive pairs (e.g., a visible image and an infrared image of the same person) within each scenario. Next, we propose a dynamic text representation update strategy to maintain consistency between text and image supervision signals. Experimental results across multiple scenarios demonstrate the superiority and generalizability of ITKM; it not only outperforms existing scenario-specific methods but also enhances overall performance by integrating knowledge from multiple scenarios.
\end{abstract}

\section{Introduction}
\label{sec:intro}
Given a query image of an individual, person re-identification (ReID) \cite{agw,caj,scarcp,gong2024cross,tan2024rle} seeks to identify images of the same person from a gallery containing a large number of person images. To reduce human labor and tedium, researchers have explored various unsupervised traditional ReID (UT-ReID) methods \cite{cap,iics,ice,cc,dccc}, which typically utilize pseudo-labels generated by clustering algorithms in place of manually annotated identity labels to guide model optimization. 
Although advanced UT-ReID methods have achieved promising performance in simple scenarios, they often struggle to handle more challenging scenarios. Therefore, researchers have explored various challenging scenarios, such as unsupervised visible-infrared ReID (UVI-ReID) \cite{chcr,shi2024multi,gur,tokenmatcher}, unsupervised clothing change ReID (UCC-ReID) \cite{cicl,japl}, and unsupervised cross-resolution ReID (UCR-ReID) \cite{drfm}, and proposed methods tailored to each scenario. 

\begin{figure}[!t]
\centering
\includegraphics[width=0.45\textwidth]{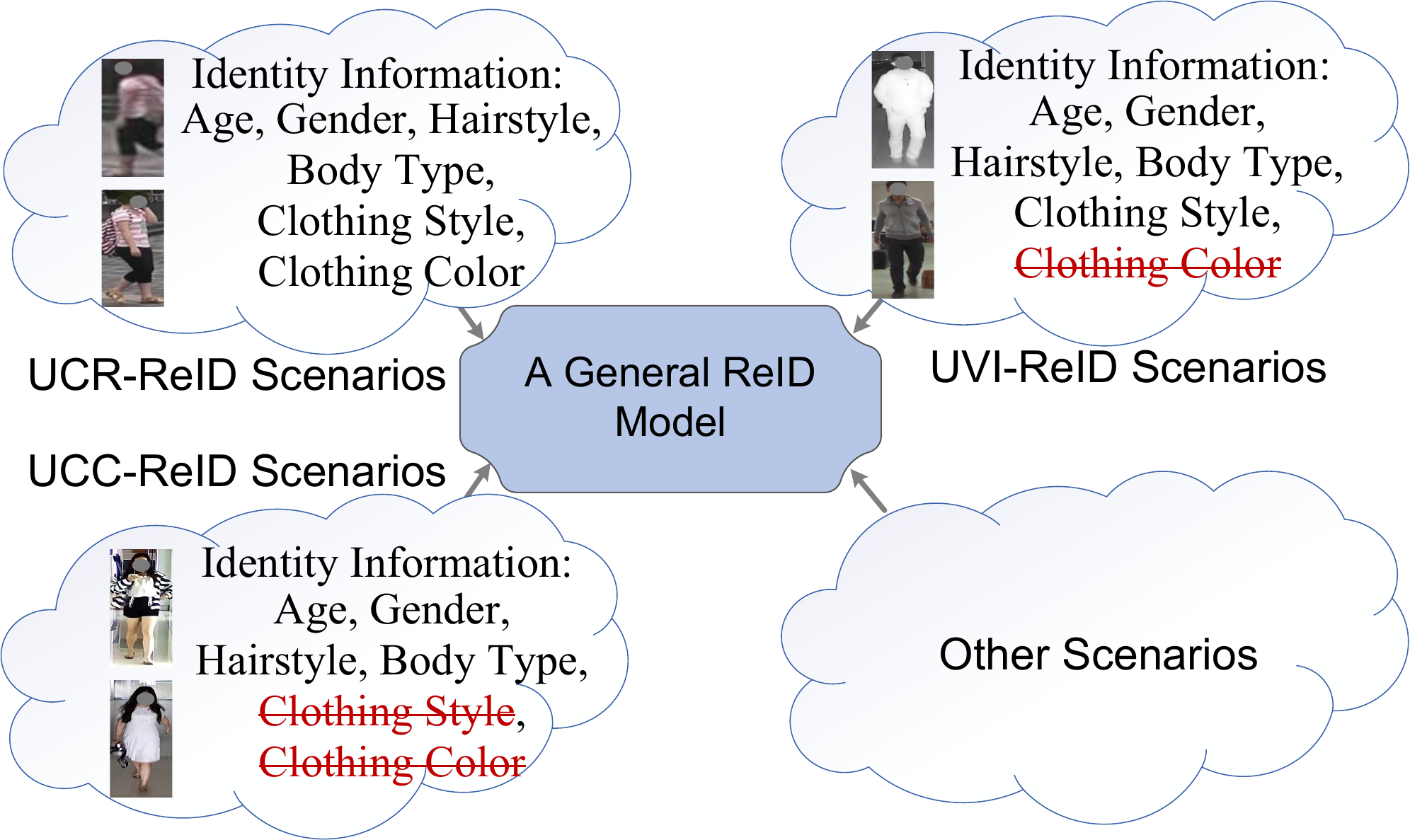}
\caption{Identity information is scenario-dependent.}
\label{f1}
\end{figure}

\par The nature of identity-related information is typically different across various scenarios. As depicted in Figure~\ref{f1}, in UCR-ReID, both the style and color of clothing are considered identity-related information \cite{drfm}. However, color is irrelevant for identity in UVI-ReID \cite{adca} because of the lack of color information in infrared images. Furthermore, UCC-ReID typically assumes that the same individual may wear different clothing, thus the style and color of clothing might not be identity-related \cite{japl}. 
Because the afore-mentioned unsupervised scenario-specific ReID (USS-ReID) methods focus on identity-related features unique to each scenario, limiting their generalizability across multiple scenarios. This results in two key limitations. 
First, practical ReID applications often span multiple scenarios—for instance, a person may change clothing or appear under low-light conditions. Existing USS-ReID methods typically require separate models for each scenario, involving distinct training strategies and multiple sets of weights, which increases both system complexity and deployment cost.
Second, existing USS-ReID methods are designed for individual scenarios and do not support joint training across multiple scenarios, thus failing to leverage the performance benefits of increased data diversity \cite{zheng2024semi,shi2023dual}.

\par To address the aforementioned issues, we propose a novel image-text knowledge modeling (ITKM) framework that effectively leverages the representational power of vision-language models (VLMs)~\cite{clip} for addressing the challenging unsupervised multi-scenario person re-identification (UMS-ReID). 
UMS-ReID is a novel task that aims to train a single general model in an unsupervised manner to handle multiple diverse scenarios—unlike prior tasks \cite{adca,japl}, which target individual scenarios. To tackle the challenges of UMS-ReID, ITKM introduces advances in both data processing and model optimization. 
In terms of data, based on the relationship between data heterogeneity (e.g., modality gaps, clothing variations, and resolution disparities) and encoder structure \cite{pgm,shi2024multi}, we consistently divide the heterogeneous data in each scenario into two homogeneous groups, ensuring that the image encoder in ITKM can accommodate inputs from all scenarios. 
In terms of optimization, given the powerful representation and transfer capabilities of VLMs, we design ITKM as a three-stage framework based on CLIP \cite{clip,coop}.
In Stage I, we introduce a scenario embedding in the CLIP image encoder and fine-tune the encoder to adapt and effectively leverage knowledge from multiple scenarios. In Stage II, to obtain scenario-specific text representations, we: 
(a) optimize a set of learned text embeddings to associate, through the CLIP, with identity pseudo-labels from Stage I 
and 
(b) introduce a multi-scenario separation loss to increase the divergence between inter-scenario text representations. 
Stage III first introduces cluster-level heterogeneous matching (CHM) and instance-level heterogeneous matching (IHM) modules to obtain reliable heterogeneous positive pairs (e.g., a visible image and an infrared image of the same person, images of the same person wearing different clothing, or images of the same person with different resolutions) within each scenario.
Next, Stage III introduces a dynamic text representation update (DRU) strategy that is guided by the latest pseudo-labels and helps maintain consistency between text and image supervision signals in unsupervised setting.

\par The main contributions are summarized as follows:
\begin{itemize}
\item[$\bullet$]We propose a novel task, UMS-ReID, along with a tailored framework, ITKM, to address it. To the best of our knowledge, this is the first study to explore unsupervised person re-identification across multiple scenarios.
\item[$\bullet$]To adapt to different scenarios and encourage scenario-specific text representations, we introduce a novel scenario embedding for the CLIP image encoder and propose a multi-scenario separation loss.
\item[$\bullet$]We construct two new modules, CHM and IHM, for obtain reliable heterogeneous positive pairs that are critical for learning in our unsupervised setting. Additionally, we introduce the DRU strategy to ensure consistency between text and image supervision signals.
\item[$\bullet$]Experimental results on datasets from different scenarios validate the superiority and generalization of ITKM, which not only surpasses existing USS-ReID methods within each scenario but also enhances overall performance by incorporating multi-scenario knowledge.
\end{itemize}

\section{Related Work}
\label{sec:rw}

UT-ReID methods \cite{pplr,purification} typically assume that different images of the same person exhibit high similarity. However, real-world applications often involve multiple challenging scenarios, where different images of the same person may exhibit significant heterogeneity. These real-world complexities are partly addressed in prior research by establishing specific tasks and tailored methods. 
For example, given a visible (infrared) image of a person, UVI-ReID \cite{sdcl} aims to find images with the same identity in a gallery composed of infrared (visible) images. UCC-ReID \cite{sicl} typically assumes that people may change clothes and seeks to match images of the same person wearing different clothing. Additionally, UCR-ReID \cite{drfm} aims to match low-resolution images with high-resolution images of the same person. 
Building on prior work, this paper proposes a novel UMS-ReID method that supports multiple scenarios simultaneously, aiming to promote broader real-world applications of ReID.

Pre-trained VLMs \cite{clip,coop,lin2024revisiting} have demonstrated promising transferability in downstream tasks. A pioneer in this area is the CLIP \cite{clip} model that trains a pair of image and text encoders by maximizing the representation similarity between matched image-text pairs. Subsequent research, such as CoOp \cite{coop}, leverages learnable text embeddings to further enhance the flexibility of CLIP. Researchers in the ReID field have also explored CLIP-based methods \cite{clipreid,cclnet,promptsg}. For example, CLIP-ReID \cite{clipreid} first learns a set of text embeddings for each identity, then uses the learned text embeddings to assist in optimizing the image encoder. 
CCLNet \cite{cclnet} optimizes learnable text embeddings based on pseudo-labels in the UVI-ReID task and uses the text representations as additional supervisory signals. The proposed ITKM framework also benefits from CLIP's representation capabilities. Unlike prior works, we focus on the UMS-ReID task and explicitly introduce DRU to update text representations, addressing the limitations of outdated offline representations used in previous methods.

\begin{figure*}[!t]
\centering
\includegraphics[width=0.95\textwidth]{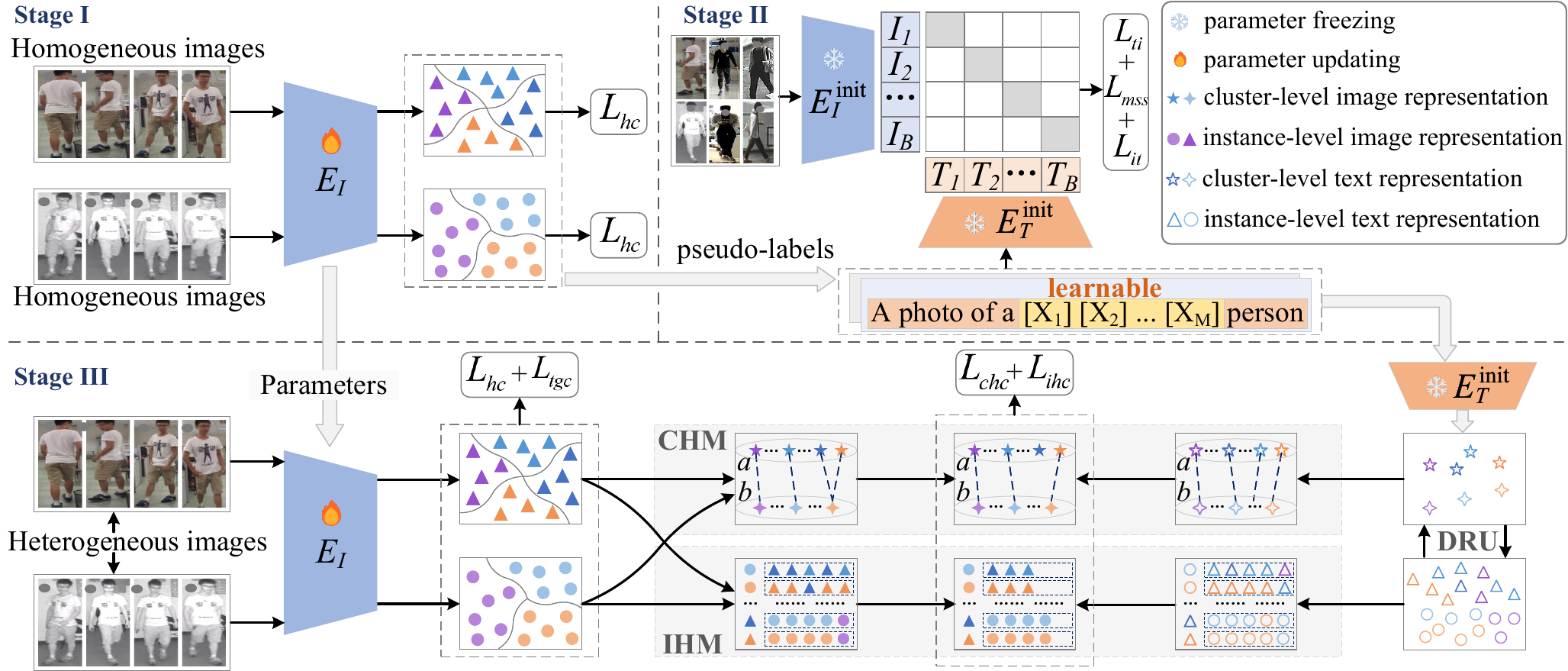}
\caption{
The proposed ITKM framework for UMS-ReID consists of three stages. Stage I performs unsupervised homogeneous learning to generate pseudo-labels. Stage II learns text embeddings in the sentence ``A photo of a ${[X_1]}{\rm{ }}[{X_2}]{\rm{ }}...{\rm{ }}[{X_M}]$ person'' to associate it with the images for a homogeneous pseudo-label via CLIP. Stage III conducts unsupervised heterogeneous learning using CHM and IHM. In Stages I and III, colors indicate pseudo-labels and shapes indicate homogeneous groups.
}
\label{f2}
\end{figure*}

\section{Proposed Framework}
To illustrate the UMS-ReID task, we use a multi-scenario setting that includes UVI-ReID \cite{sdcl}, UCC-ReID \cite{cicl}, and UCR-ReID \cite{drfm}. During the training phase, we define the combination of the training sets from these three scenarios as $\{ {X^s}\} _{s = 1}^S$, where $s$ indexes the $S{\rm{ = }}3$ scenarios. This paper aims to demonstrate the potential of UMS-ReID. Accordingly, we do not prioritize large-scale datasets or extensive model parameters, leaving these aspects for future work.
Given the data distribution, we first divide the heterogeneous images within each scenario into two homogeneous groups, labeled as $a$ and $b$. Further details are provided in the Supplementary Materials, where the complete ITKM procedure is also summarized as Algorithm~\ref{alg1}.
As shown in Figure \ref{f2}, our proposed ITKM framework consists of three stages: \emph{unsupervised homogeneous learning} (Stage I), \emph{text representation learning} (Stage II), and \emph{unsupervised heterogeneous learning}  (Stage III).

\subsection{Unsupervised Homogeneous Learning}
\begin{figure}[!t]
\centering
\includegraphics[width=0.475\textwidth]{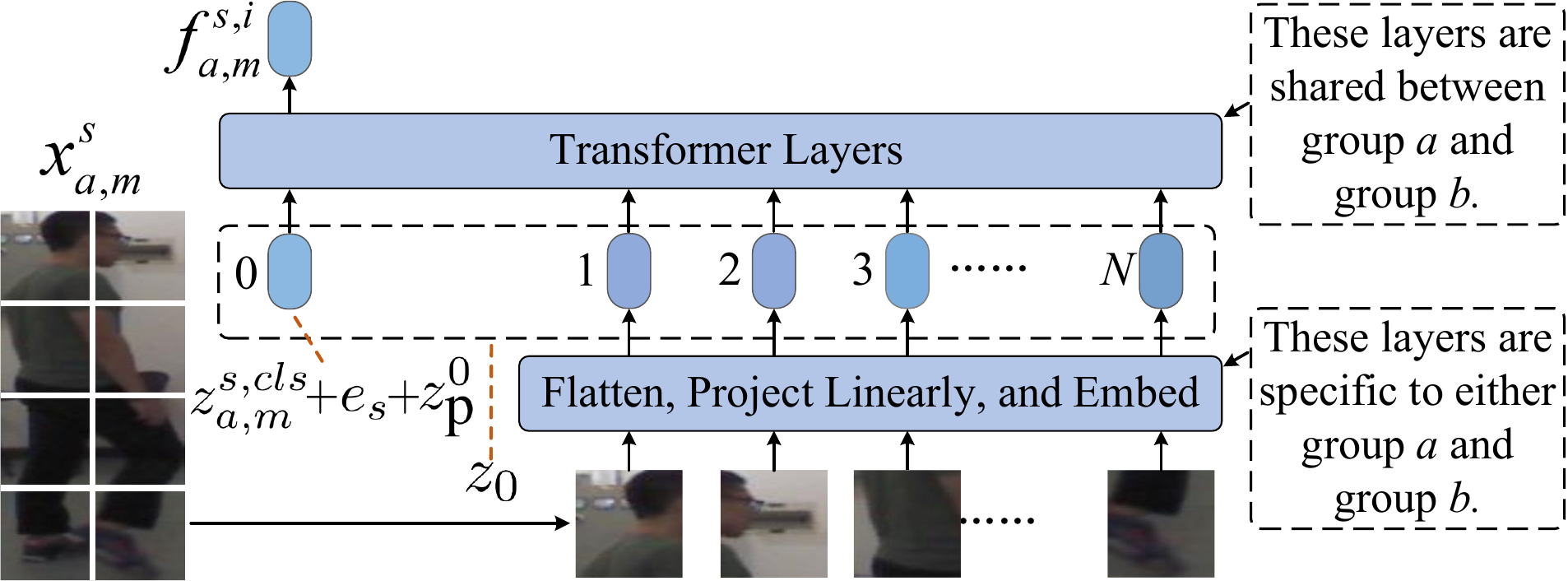}
\caption{Architecture of image encoder used to obtain identity representations $f_{a,m}^{s,i}$ from an input image $x_{a,m}^s$.}
\label{f3}
\end{figure}

In Stage I, we adapt the image encoder from a pre-trained CLIP model to transform an input image into a scenario-specific representation that preserves identity information. Specifically, we adapt the vision transformer (ViT)~\cite{Dosovitskiy:VIT:ICLR2021} $E_I^{\textrm{init}}$, which constitutes the CLIP image encoder, to introduce a dual-branch frontend, where each branch is used for one homogeneous group, and, additionally, introduce a scenario embedding in the image encoder. The process is illustrated in Figure \ref{f3} using an input image $x_{a,m}^s$ from the group $a$ of $s$-th scenario. The image  $x_{a,m}^s$ is partitioned into $N$ patches which are flattened into vectors $x_1, x_2, \ldots  x_N$ and a learnable embedding matrix $P$ maps each of the patches to a $1 \times D$ vector. The class embedding is concatenated at the head of the sequence of vectors and position and scenario are then additively added into the resulting vector to obtain an $(N+1)$-length sequence of $1 \times D$ vectors
\begin{equation}
\label{eq:LinPosNScenEmbed}
{z_0} = \left[ {z_{a,m}^{s,cls} + {e_s};[ x_1 ; x_2; \cdots ; x_N ] P} \right] + [ z_{\textrm{p}}^0;\cdots ;z_{\textrm{p}}^N ]
\end{equation}
where $[ z_{\textrm{p}}^0;\cdots ;z_{\textrm{p}}^N]$ represent the position embeddings and $e_s$ is the learnable scenario embedding. The flattening and projection of the patches, concatenation of $z_{a,m}^{s,cls}$, and the position embedding are based on the standard ViT approach~\cite{Dosovitskiy:VIT:ICLR2021}. The scenario embedding is introduced in this work to address our UMS-ReID setting.

As shown in Figure \ref{f3}, $z_0$ is processed through a series of transformer layers of the ViT, producing (at the head of the final layer) the identity representation $f_{a,m}^{s,i}$, which is referred to as the instance-level image representation in the following description. The instance-level image representation $f_{b,m}^{s,i}$ for group $b$ is similarly obtained. 
The instance-level image representations are then clustered within each homogeneous group to obtain homogeneous pseudo-labels. 

\par We then calculate cluster-level image representations. For example, the representation of the $u$-th cluster in group $a$ in the $s$-th scenario is defined as:
\begin{equation}
\label{eq2}
c_{a,u}^{s,i} = \frac{1}{{N_{a,u}^s}}\sum\limits_{m = 1}^{N_{a,u}^s} {f_{a,m}^{s,i}} \textrm{,}
\end{equation}
where $N_{a,u}^s$ represents the number of images in this cluster and $f_{a,m}^{s,i}$ denotes the instance-level image representation in this cluster. 
Subsequently, we refine the image encoder $E_I^{\textrm{init}}$ based on homogeneous learning, and the optimized image encoder is referred to as $E_I$. 
Specifically, the homogeneous contrastive loss is introduced to optimize $E_I$ based on the pseudo-labels. For an instance-level image representation $f_{a,m}^{s,i}$, the homogeneous contrastive loss is defined as:
\begin{equation}
\label{eq4}
L_{hc}^{s,a,m} =  - \log \frac{{\exp (f_{a,m}^{s,i} \cdot c{{_{a,u}^{s,i}}^T}/{\tau})}}{{\sum\nolimits_{v = 1}^{C_a^{s,i}} {\exp (f_{a,m}^{s,i} \cdot c{{_{a,v}^{s,i}}^T}/{\tau})} }} \textrm{,}
\end{equation}
where $c_{a,u}^{s,i}$ is from the same cluster as $f_{a,m}^{s,i}$, $C_a^{s,i}$ represents the number of clusters in group $a$ in the $s$-th scenario, and ${\tau}$ is the temperature hyperparameter. For an instance-level representation $f_{b,m}^{s,i}$ in group $b$, the corresponding loss $L_{hc}^{s,b,m}$ is obtained analogously. The overall homogeneous contrastive loss is then obtained as:
\begin{equation}
\label{eq6}
{L_{hc}} = \frac{1}{{B \cdot S}}\sum\limits_{s = 1}^S {\sum\limits_{m = 1}^B {(L_{hc}^{s,a,m} + L_{hc}^{s,b,m})} } \textrm{,}
\end{equation}
where $B$ is the batch size. Clustering for assigning homogeneous pseudo-labels and optimizing $E_I$ (to minimize the loss in Eq.~\ref{eq6}) are performed iteratively. The final pseudo-labels are then used in Stage II.

\subsection{Text Representation Learning}
Stage II, which is illustrated in Figure \ref{f2}, learns identity related text representations by using a (second) pre-trained CLIP model with an image encoder $E_I^{\textrm{init}}$ and a text encoder $E_T^{\textrm{init}}$. Specifically, for each homogeneous pseudo-label saved during Stage I, a set of text embeddings ${[X]_1}{\rm{ }}[{X_2}]{\rm{ }}...{\rm{ }}[{X_M}]$ are learned to optimize the association between the \emph{pseudo-label tagging sentence}  ``A photo of a ${[X]_1}{\rm{ }}[{X_2}]{\rm{ }}...{\rm{ }}[{X_M}]$ person'' and the images with the pseudo-label by minimizing an overall loss:
\begin{equation}
\label{eq7}
{L_{s2}} =  \frac{1}{{B \cdot S}} \sum\limits_{s = 1}^S \sum\limits_{m = 1}^B \left ( {L_{it}^{s,m}} + {L_{ti}^{s,m}} \right ) + {\lambda _{mss}}{L_{mss}} \textrm{,}
\end{equation}
where $L_{it}^{s,m}$ and $L_{ti}^{s,m}$ are image-to-text and text-to-image contrastive losses, respectively, $L_{mss}$ is a multi-scenario separation loss designed to encourage scenario-specific text representations and enable the model to effectively adapt to different scenarios, and ${\lambda _{mss}}$ is a weight hyperparameter that controls the relative significance of  $L_{mss}$. These individual components of the overall loss are defined as:
\begin{equation}
L_{it}^{s,m} =  - {\log \frac{{\exp (f_m^{s,i} \cdot f{{_m^{s,t}}^T})}}{{\sum\nolimits_{v = 1}^B {\exp (f_m^{s,i} \cdot f{{_v^{s,t}}^T})} }}}  \textrm{,} \label{eq8}
\end{equation}
\begin{equation}
L_{ti}^{s,m} =  {\frac{-1}{{|\varphi_m^{s,t}|}} \! \sum\limits_{f_u^{s,i} \in \varphi_m^{s,t}} \!\!{\log \frac{{\exp (f_m^{s,t} \cdot f{{_u^{s,i}}^T})}}{{\sum\nolimits_{v = 1}^B {\exp (f_m^{s,t} \cdot f{{_v^{s,i}}^T})} }}} } \textrm{,} \label{eq9}
\end{equation}
\begin{equation}
{L_{mss}} =  \sum\limits_{g = 1}^S {\sum\limits_{\scriptstyle h = 1\hfill\atop \scriptstyle h \ne g\hfill}^S {{{\left[ {\kappa - \left\| {\frac{1}{B}\sum\limits_{m = 1}^B {(f_m^{g,t}}  - f_m^{h,t})} \right\|_2^2} \right]}_ + }} } \textrm{,} \label{eq10}                           
\end{equation}
where $f_v^{s,i}$ denotes the representation of the $v$-th image in the batch, $f_v^{s,t}$  is the representation of the sentence for the pseudo-label assigned to the $v$-th image, $\varphi_m^{s,t}$ denotes the set of image representations (in the batch) that share the same pseudo-label as the $f_m^{s,t}$, $\kappa$ is a margin hyperparameter that ensures adequate separation between inter-scenario text representations, and $[\zeta]_{+}$ denotes the positive part of (the real-number) $\zeta$. Note that the learnable text embeddings ${[X]_1}{\rm{ }}[{X_2}]{\rm{ }}...{\rm{ }}[{X_M}]$ are optimized by minimizing the loss in Eq.~\ref{eq7} while keeping the parameters of the encoders $E_I^{\textrm{init}}$ and $E_T^{\textrm{init}}$ frozen. Since the same operations are performed on all images and do not involve interactions between heterogeneous images, we do not differentiate between them in the equations.

We refer to the text representation output by $ E_T^{\textrm{init}}$ for a sentence as the cluster-level text representation. The final cluster-level text representations for the sentences are saved, and for each image within a cluster, an instance-level text representation is initialized as the corresponding cluster-level text representation. The saved text representations are referred to as \emph{offline text representations}.

\subsection{Unsupervised Heterogeneous Learning}
Stage III aims to optimize the dual front-end image encoder from Stage I to ensure its identity representations are effective for UMS-ReID. To achieve this, we employ an iterative approach that alternates between updating the image encoder and leveraging the current text representations.
Specifically, each iteration in Stage III begins by using the current image encoder $E_I$ to obtain instance-level image representations for $\{ {X^s}\} _{s = 1}^S$. These representations are then clustered within each homogeneous group to generate homogeneous pseudo-labels and cluster-level image representations are computed similarly to Eq.~\ref{eq2}. 
DRU is utilized to obtain instance-level \emph{online text representations} guided by the latest pseudo-labels. 
CHM and IHM are the utilized to obtain cluster-level heterogeneous image pairs and instance-level heterogeneous positive sets, respectively.
An overall contrastive loss—comprising homogeneous, cluster-level heterogeneous, instance-level heterogeneous, and text-guided contrastive losses—is used to update $E_I$. 
Details of DRU, CHM and IHM are provided next.

\subsubsection{Dynamic text representation update.}
The offline text representations saved in Stage II may have pseudo-labels that are inconsistent with the newly obtained pseudo-labels in Stage III, potentially misleading the optimization process. To address this, we develop a dynamic text representation update (DRU) strategy to obtain \emph{online text representations}. Mutual updating between cluster-level and instance-level text representations is accomplished through DRU. Specifically, after each clustering iteration in Stage III, we first calculate the cluster-level text pseudo-representation:
\begin{equation}
\label{eq17}
\hat c_u^t = \frac{1}{{N_u^s}}\sum\limits_{m = 1}^{N_u^s} {f_m^{s,t}} \textrm{,}
\end{equation}
where $N_u^s$ is the number of images in the $u$-th cluster of the $s$-th scenario, and $f_m^{s,t}$ is the instance-level text representation in the corresponding cluster.
Subsequently, we define the top $\eta $ proportion of instance-level text representations within the cluster that are closest to $\hat c_u^t$ as the neighboring representation set $\xi (\hat c_u^t)$, and then calculate the new cluster-level text representation:
\begin{equation}
\label{eq18}
c_u^t = \frac{1}{{|\xi (\hat c_u^t)|}}\sum\limits_{f_m^{s,t} \in \xi (\hat c_u^t)} {f_m^{s,t}} \textrm{.}
\end{equation}
Then, based on the new cluster-level text representation, we update each instance-level text representation:
\begin{equation}
\label{eq19}
f_m^{s,t} \leftarrow (1 - \alpha )f_m^{s,t} + \alpha c_u^t \textrm{,}
\end{equation}
where $\alpha$ is the update rate hyperparameter. From the above process, it is evident that the DRU allows text representations to be updated alongside pseudo-labels.

DRU effectively prevents inconsistencies between the text and image supervision signals. 
Moreover, this strategy has the potential to ensure inter-cluster separability and intra-cluster compactness. 
On one hand, based on Eq. \ref{eq17} and Eq. \ref{eq18}, DRU only utilizes the inner instance-level text representations to compute the new cluster-level text representation, thereby ensuring inter-cluster separability. On the other hand, DRU updates all instance-level text representations within the cluster using the new cluster-level text representation, enhancing intra-cluster compactness.
\par Finally, we design a text-guided contrastive loss to provide supervisory signals from text semantics for $E_I$. For any $f_m^{s,i}$, the text-guided contrastive loss is defined as:
\begin{equation}
\label{eq20}
{L_{tgc}} =  - \frac{1}{{B \cdot S}}\sum\limits_{s = 1}^S {\sum\limits_{m = 1}^B {\log \frac{{\exp (f_m^{s,i} \cdot f{{_m^{s,t}}^T})}}{{\sum\nolimits_{v = 1}^B {\exp (f_m^{s,i} \cdot f{{_v^{s,t}}^T})} }}} } \textrm{,}
\end{equation}
where $f_m^{s,i}$ and $f_m^{s,t}$ are the instance-level image and text representations of the same image, respectively.

\subsubsection{Cluster-level heterogeneous matching.}
CHM aims to identify cluster-level heterogeneous image pairs of the same identity, with each pair consisting of one cluster from each of the two homogeneous groups.  
After each clustering, based on cluster-level image representations, we match cluster-level heterogeneous image pairs and cluster-level heterogeneous text pairs using the graph matching strategy \cite{pgm}. For example, in UVI-ReID, a cluster-level heterogeneous image pair consists of one cluster composed of infrared images and another composed of visible images, both presumed to contain images of the same person. We next assess consistency between cluster-level heterogeneous image and text pairs. For consistent pairs, we directly retain the corresponding cluster-level heterogeneous image pairs; for each inconsistent pair, we stochastically retain the corresponding cluster-level heterogeneous image pair with a probability $\beta \in [0,1]$. Each retained cluster-level heterogeneous image pair corresponds to two clusters that are (putatively) mutually heterogeneous positive clusters. Finally, we construct a cluster-level heterogeneous contrastive loss to mitigate intra-scenario heterogeneity. For any $f_{a,m}^{s,i}$, the cluster-level heterogeneous contrastive loss is defined as:
\begin{equation}
\label{eq11}
L_{chc}^{s,ab,m} =  - \log \frac{{\exp (f_{a,m}^{s,i} \cdot c{{_{b,u}^{s,i}}^T}/{\tau})}}{{\sum\nolimits_{v = 1}^{C_b^{s,i}} {\exp (f_{a,m}^{s,i} \cdot c{{_{b,v}^{s,i}}^T}/{\tau})} }} \textrm{,}
\end{equation}
where $c_{b,u}^{s,i}$ represents the cluster-level image representation of the heterogeneous positive cluster corresponding to $f_{a,m}^{s,i}$, $C_b^{s,i}$ represents the number of clusters in group $b$ in the $s$-th scenario. For any $f_{b,m}^{s,i}$, the cluster-level heterogeneous contrastive loss $L_{chc}^{s,ba,m}$ is similarly defined. The overall cluster-level heterogeneous contrastive loss is then defined as:
\begin{equation}
\label{eq13}
{L_{chc}} = \frac{1}{{B \cdot S}}\sum\limits_{s = 1}^S {\sum\limits_{m = 1}^B {(L_{chc}^{s,ab,m} + L_{chc}^{s,ba,m})} } \textrm{.}
\end{equation}

\subsubsection{Instance-level heterogeneous matching.} 
In addition to matching heterogeneous positive clusters, we also focus on obtaining (putative) heterogeneous positive sets at the instance-level. Specifically, we first search for heterogeneous neighbor sets for each image in both image and text representation spaces based on cosine similarity. For the instance-level image representation $f_{a,m}^{s,i}$ of any given image $x_{a,m}^s$, we define the heterogeneous neighbor set ${\psi_i}(x_{a,m}^s)$ as the set of images in group $b$ corresponding to the top-$k$ instance-level image representations that have the highest cosine similarity with $f_{a,m}^{s,i}$. Similarly, we define the heterogeneous neighbor set ${\psi_t}(x_{a,m}^s)$ as the set of images in group $b$ corresponding to the top-$k$ instance-level text representations that have the highest cosine similarity with the instance-level text representation $f_{a,m}^{s,t}$ of the image $x_{a,m}^s$. Subsequently, we take the set of the instance-level image representations corresponding to all images in the intersection ${\psi_i}(x_{a,m}^s) {\rm{\cap}} {\psi_t}(x_{a,m}^s) $ as the instance-level heterogeneous positive set $U_{a,m}^{s,i} {\rm{ = }}  \{ f_{b,u}^{s,i}\}$ for $f_{a,m}^{s,i}$. Similarly, we also identify a instance-level heterogeneous positive set $U_{b,m}^{s,i} {\rm{ = }} \{ f_{a,u}^{s,i}\}$ for $f_{b,m}^{s,i}$. Finally, we construct an instance-level heterogeneous contrastive loss to further bridge intra-scenario heterogeneity. For $f_{a,m}^{s,i}$, the instance-level heterogeneous contrastive loss is defined as:
\begin{equation}
\label{eq14}
L_{ihc}^{s,ab,m} \! =  \!  \frac{-1}{{{\rm{|}}U_{a,m}^{s,i}{\rm{|}}}} \! \! \sum\limits_{f_{b,u}^{s,i} \in U_{a,m}^{s,i}} \!\!\!\! {\log \frac{{\exp (f_{a,m}^{s,i} \cdot f{{_{b,u}^{s,i}}^T}/{\tau})}}{{\sum\nolimits_{v = 1}^{N_b^s} {\exp (f_{a,m}^{s,i} \cdot f{{_{b,v}^{s,i}}^T}/{\tau})} }}} \textrm{,}
\end{equation}
where $N_b^s$ represents the number of images in group $b$ within the $s$-th scenario. For $f_{b,m}^{s,i}$, the loss $L_{ihc}^{s,ba,m}$ is similarly defined. The overall instance-level heterogeneous contrastive loss is then obtained as:
\begin{equation}
\label{eq16}
{L_{ihc}} = \frac{1}{{B \cdot S}}\sum\limits_{s = 1}^S {\sum\limits_{m = 1}^B {(L_{ihc}^{s,ab,m} + L_{ihc}^{s,ba,m})} } \textrm{.}
\end{equation}
\par For any given image, we refer to the combination of the image with any image from the heterogeneous positive cluster obtained by CHM, as well as the combination with any image from the instance-level heterogeneous positive set obtained by IHM, as a \emph{heterogeneous positive pair}. 

\par In summary, the overall objective function of Stage III is defined as:
\begin{equation}
\label{eq21}
{L_{s3}} = {L_{hc}} + {L_{chc}} + {L_{ihc}} + {\lambda _{tgc}}{L_{tgc}} \textrm{,}
\end{equation}
where ${\lambda _{tgc}}$ is a weight hyperparameter for ${L_{tgc}}$. The final trained image encoder $E_I$ for ITKM is obtained by optimizing ${L_{s3}}$.

The complete ITKM procedure is summarized as Algorithm 1 in the Supplementary Materials.

\section{Experiments}
\subsection{Datasets and Evaluation Metrics}
We evaluate the proposed method on the commonly used SYSU-MM01 \cite{sysumm01}, LTCC \cite{ltcc}, and MLR-CUHK03 \cite{drfm}, using mean average precision (mAP) and cumulative matching characteristic (CMC) as evaluation metrics. The Supplementary Materials include dataset and implementation details.

\begin{table*}
  \centering 
{\fontsize{9}{10}\selectfont
\begin{tabular}{c|cccc|cccc|ccc}
\hline
\multirow{3}{*}{Methods} & \multicolumn{4}{c|}{SYSU-MM01}                                                            & \multicolumn{4}{c|}{LTCC}                                                          & \multicolumn{3}{c}{\multirow{2}{*}{MLR-CUHK03}} \\ \cline{2-9}
                         & \multicolumn{2}{c|}{All Search}                    & \multicolumn{2}{c|}{Indoor   Search} & \multicolumn{2}{c|}{Clothing   Change}             & \multicolumn{2}{c|}{General}  & \multicolumn{3}{c}{}                            \\ \cline{2-12} 
                         & Rank-1        & \multicolumn{1}{c|}{mAP}           & Rank-1            & mAP              & Rank-1        & \multicolumn{1}{c|}{mAP}           & Rank-1        & mAP           & Rank-1         & Rank-5         & Rank-10       \\ \hline
ICE                      & 20.5          & \multicolumn{1}{c|}{20.4}          & 29.8              & 38.4             & 14.5          & \multicolumn{1}{c|}{7.1}           & \textbf{61.1} & \textbf{28.4} & 27.6           & 65.3           & 80.6          \\
CC                       & 20.2          & \multicolumn{1}{c|}{\textbf{22.0}} & 23.3              & 34.0             & 7.4           & \multicolumn{1}{c|}{6.0}           & 17.0          & 11.0          & \textbf{31.6}  & \textbf{66.8}  & \textbf{80.8} \\
Purification             & \textbf{20.8} & \multicolumn{1}{c|}{21.3}          & \textbf{30.6}     & \textbf{39.0}    & \textbf{15.6} & \multicolumn{1}{c|}{\textbf{8.7}}  & 52.8          & 23.9          & 30.5           & 65.8           & 78.9          \\
DCCC                     & 18.8          & \multicolumn{1}{c|}{18.6}          & 21.3              & 33.1             & 13.7          & \multicolumn{1}{c|}{7.2}           & 47.6          & 21.3          & 27.6           & 60.3           & 73.6          \\ \hline
SDCL                     & 64.5          & \multicolumn{1}{c|}{\textbf{63.2}} & 71.4              & \textbf{76.9}    & -             & \multicolumn{1}{c|}{-}             & -             & -             & -              & -              & -             \\
TokenMatcher             & 65.1          & \multicolumn{1}{c|}{62.8}          & 69.0              & 74.9             & -             & \multicolumn{1}{c|}{-}             & -             & -             & -              & -              & -             \\
CICL                     & -             & \multicolumn{1}{c|}{-}             & -                 & -                & 25.7          & \multicolumn{1}{c|}{13.0}          & 60.3          & 29.6          & -              & -              & -             \\
JAPL                     & -             & \multicolumn{1}{c|}{-}             & -                 & -                & 26.1          & \multicolumn{1}{c|}{12.9}          & \textbf{63.0} & \textbf{31.8} & -              & -              & -             \\
DRFM                     & -             & \multicolumn{1}{c|}{-}             & -                 & -                & -             & \multicolumn{1}{c|}{-}             & -             & -             & 35.8           & 72.2           & 83.6          \\
ITKM(S)                  & 64.6          & \multicolumn{1}{c|}{\textbf{63.2}} & 72.0              & 76.7             & \textbf{26.2} & \multicolumn{1}{c|}{\textbf{13.2}} & 62.8          & 31.6          & \textbf{62.5}  & \textbf{85.1}  & \textbf{90.3} \\ \hline
SDCL(M)                  & 63.0          & \multicolumn{1}{c|}{61.8}          & 69.2              & 75.0             & 18.8          & \multicolumn{1}{c|}{9.3}           & 55.2          & 25.7          & 35.3           & 71.6           & 82.1          \\
TokenMatcher(M)          & 62.5          & \multicolumn{1}{c|}{60.9}          & 68.7              & 73.6             & 16.9          & \multicolumn{1}{c|}{9.0}           & 55.3          & 25.5          & 37.7           & 72.5           & 83.0          \\
ITKM(M)                  & \textbf{64.9} & \multicolumn{1}{c|}{\textbf{63.3}} & \textbf{72.3}     & \textbf{77.1}    & \textbf{27.3} & \multicolumn{1}{c|}{\textbf{14.4}} & \textbf{63.3} & \textbf{32.5} & \textbf{63.6}  & \textbf{85.7}  & \textbf{91.6} \\ \hline
\end{tabular}}
\caption{Comparison of ITKM with existing methods on SYSU-MM01, LTCC and MLR-CUHK03.\label{t1}}
\end{table*}

\subsection{Comparison with State-of-the-Art}
In Table \ref{t1}, we evaluate ITKM trained in a single scenario (denoted as ITKM(S)) and ITKM jointly trained across three scenarios (denoted as ITKM(M)) on SYSU-MM01, LTCC, and MLR-CUHK03, respectively. 

First, we find that ITKM(S) outperforms existing unsupervised traditional (UT) methods, including ICE \cite{ice}, CC \cite{cc}, Purification \cite{purification}, and DCCC \cite{dccc}. 
This is because UT-ReID does not account for the heterogeneity within scenarios, making it difficult to obtain sufficient heterogeneous positive pairs, whereas ITKM utilizes CHM and IHM to acquire ample heterogeneous positive pairs. 

Next, we compare ITKM with unsupervised scenario-specific (USS) ReID methods for each individual scenario, including the UVI-ReID methods SDCL \cite{sdcl} and TokenMatcher \cite{tokenmatcher}, the UCC-ReID methods CICL \cite{cicl} and JAPL \cite{japl}, and the UCR-ReID method DRFM \cite{drfm}. 
We find that ITKM(S) is competitive with state-of-the-art USS-ReID methods on SYSU-MM01 and LTCC, and significantly outperforms the USS-ReID method DRFM on MLR-CUHK03. These results not only validate the superiority of ITKM but also confirm its generalizability across multiple scenarios. 

Finally, we train and test two existing methods, SDCL~\cite{sdcl} and TokenMatcher~\cite{tokenmatcher}, on the unsupervised multi-scenario (UMS) setting that combines the three datasets. These methods are denoted as TokenMatcher(M) and SDCL(M), respectively. As shown in Table \ref{t1}, joint training across multiple scenarios does not improve the performance of TokenMatcher(M) and SDCL(M); instead, they perform worse compared to single-scenario training. This indicates that these methods struggle to adapt to multiple scenarios. In contrast, ITKM(M), which is trained across multiple scenarios, shows slight improvements in performance over ITKM(S) across all three scenarios and significantly outperforms TokenMatcher(M) and SDCL(M). 
This is because the scenario embedding and multi-scenario separation loss in ITKM enable the image encoder to adapt to different scenarios, thereby improving the model's generalization ability. Further comparison is presented in the Supplementary Materials.

\begin{table*}
  \centering 
{\fontsize{9}{10}\selectfont
\begin{tabular}{c|ccccc|cc|cc|cc}
\hline
\multirow{2}{*}{Methods} & \multirow{2}{*}{$e^s$} & \multirow{2}{*}{${L_{mss}}$} & \multirow{2}{*}{DRU} & \multirow{2}{*}{Cluster} & \multirow{2}{*}{Instance} & \multicolumn{2}{c|}{SYSU-MM01(All   Search)} & \multicolumn{2}{c|}{LTCC(Clothing Change)} & \multicolumn{2}{c}{MLR-CUHK03} \\ \cline{7-12} 
                         &                        &                              &                      &                                &                                 & Rank-1                 & mAP                 & Rank-1                & mAP                & Rank-1         & Rank-5        \\ \hline
M1                       &                        &                              &                      & PGM                            &                                 & 53.8                   & 49.0                & 15.2                  & 9.7                & 33.2           & 68.6          \\
M2                       & \checkmark             &                              &                      & PGM                            &                                 & 56.1                   & 52.2                & 17.6                  & 10.7               & 34.4           & 71.2          \\
M3                       & \checkmark             & \checkmark                   &                      & PGM                            &                                 & 57.0                   & 52.7                & 18.5                  & 10.9               & 36.3           & 72.5          \\
M4                       & \checkmark             & \checkmark                   & \checkmark           & PGM                            &                                 & 59.3                   & 54.9                & 20.2                  & 11.3               & 37.0           & 72.9          \\
M5                       & \checkmark             & \checkmark                   & \checkmark           & CHM                            &                                 & 62.6                   & 59.1                & 23.8                  & 12.3               & 50.7           & 79.8          \\
M6                       & \checkmark             & \checkmark                   & \checkmark           & CHM                            & CNL                             & 63.2                   & 61.5                & 25.6                  & 13.5               & 56.5           & 82.1          \\
M7/ITKM               & \checkmark             & \checkmark                   & \checkmark           & CHM                            & IHM                             & 64.9                   & 63.3                & 27.3                  & 14.4               & 63.6           & 85.7          \\ \hline
\end{tabular}}
\caption{Results of the ablation study. “Cluster” and “Instance” refer to the methods used for obtaining heterogeneous positive clusters and instance-level heterogeneous positive sets, respectively.\label{t5}}
\end{table*}

\subsection{Ablation Study}
In Table~\ref{t5}, we assess the effectiveness of the five components within ITKM, specifically: scenario embedding ${e^s}$, multi-scenario separation loss ${L_{mss}}$, DRU, CHM, and IHM. M1 uses PGM \cite{pgm} to obtain heterogeneous positive pairs and employs ACCL \cite{pgm} along with offline text representations to jointly optimize the image encoder. Other ablation methods sequentially embed the above components or alternatives into M1.

\subsubsection{Effectiveness of ${e^s}$, ${L_{mss}}$ and DRU.} 
\begin{figure}[!t]
    \centering
    \begin{subfigure}[b]{0.225\textwidth}
        \centering
        \includegraphics[width=\textwidth]{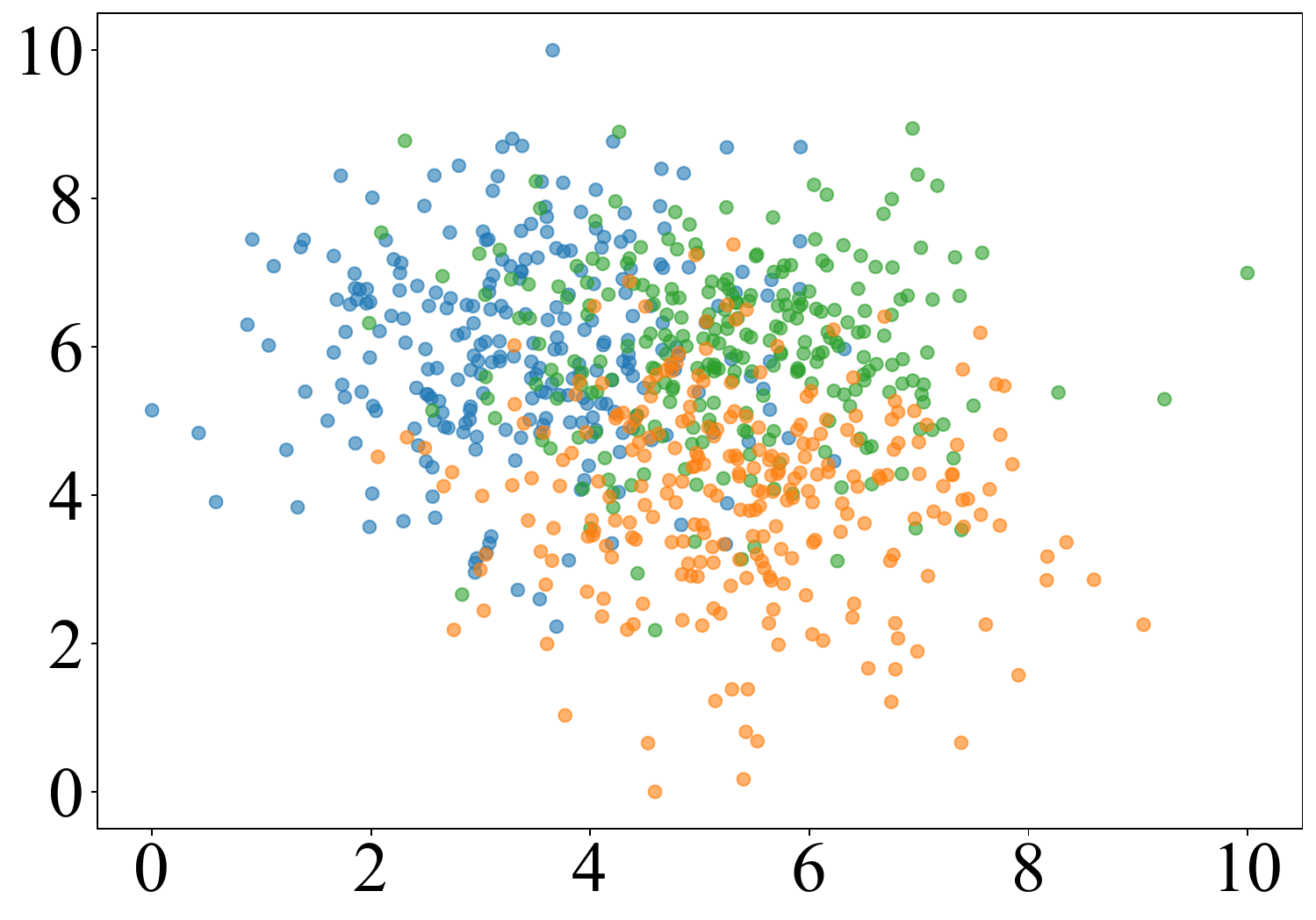}
        \caption{M2}
        \label{f4a}
    \end{subfigure}
    \hfill
    \begin{subfigure}[b]{0.225\textwidth}
        \centering
        \includegraphics[width=\textwidth]{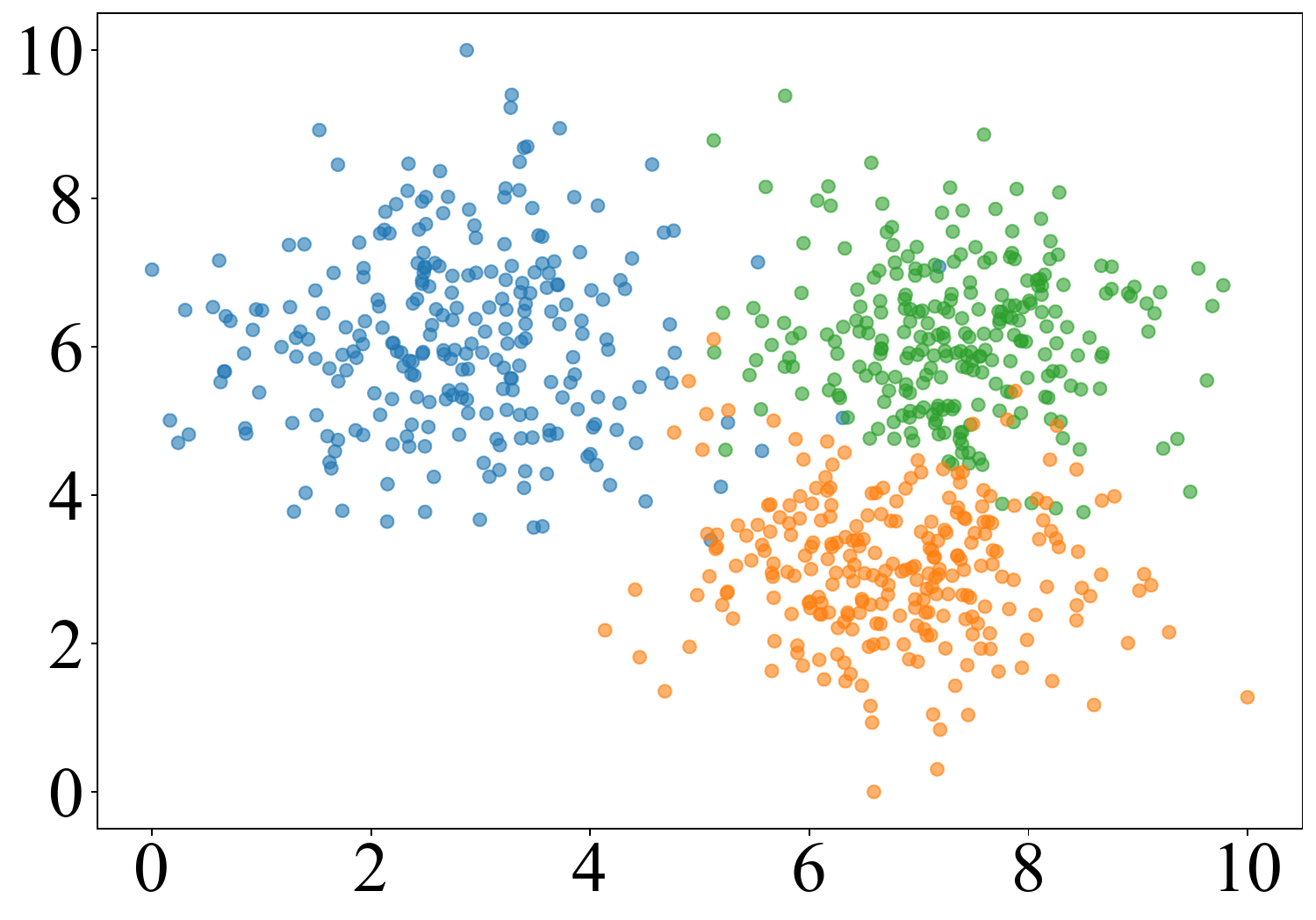}
        \caption{M3}
        \label{f4b}
    \end{subfigure}
    \caption{Two-dimensional t-SNE \cite{tsne} visualizations of the text representations for ablated methods M2 and M3.  Different colors representing text representations from different scenarios.}
    \label{f4}
\end{figure}

As shown in Table \ref{t5}, M2 achieves a significant performance improvement over M1 after introducing scenario embedding ${e^s}$. Specifically, the Rank-1 accuracy increases by 2.3\%, 2.4\%, and 1.2\% on SYSU-MM01, LTCC, and MLR-CUHK03, respectively. This validates the effectiveness of scenario embedding in the UMS-ReID task. Additionally, M3 further improves recognition performance by introducing ${L_{mss}}$ on top of M2. To further evaluate the effectiveness of ${L_{mss}}$, we use t-SNE \cite{tsne} visualizations of the text representations corresponding to the text embeddings learned by M2 and M3 during Stage II. To avoid the influence of data heterogeneity, we only visualize the text representations corresponding to high-resolution visible images, excluding those corresponding to infrared or low-resolution images. As shown in Figure \ref{f4}, compared to M2, the text representations of M3 exhibit greater inter-scenario separability. This confirms that ${L_{mss}}$ enhances model performance by encouraging scenario-specific text representations. As shown in Table \ref{t5}, M4 achieves Rank-1 accuracy improvements of 2.3\%, 1.7\%, and 0.7\% over M3 on SYSU-MM01, LTCC, and MLR-CUHK03, respectively. These results validate the effectiveness of DRU.

\subsubsection{Effectiveness of CHM.} 
\begin{figure}[!t]
    \centering
    \begin{subfigure}[b]{0.225\textwidth}
        \centering
        \includegraphics[width=\textwidth]{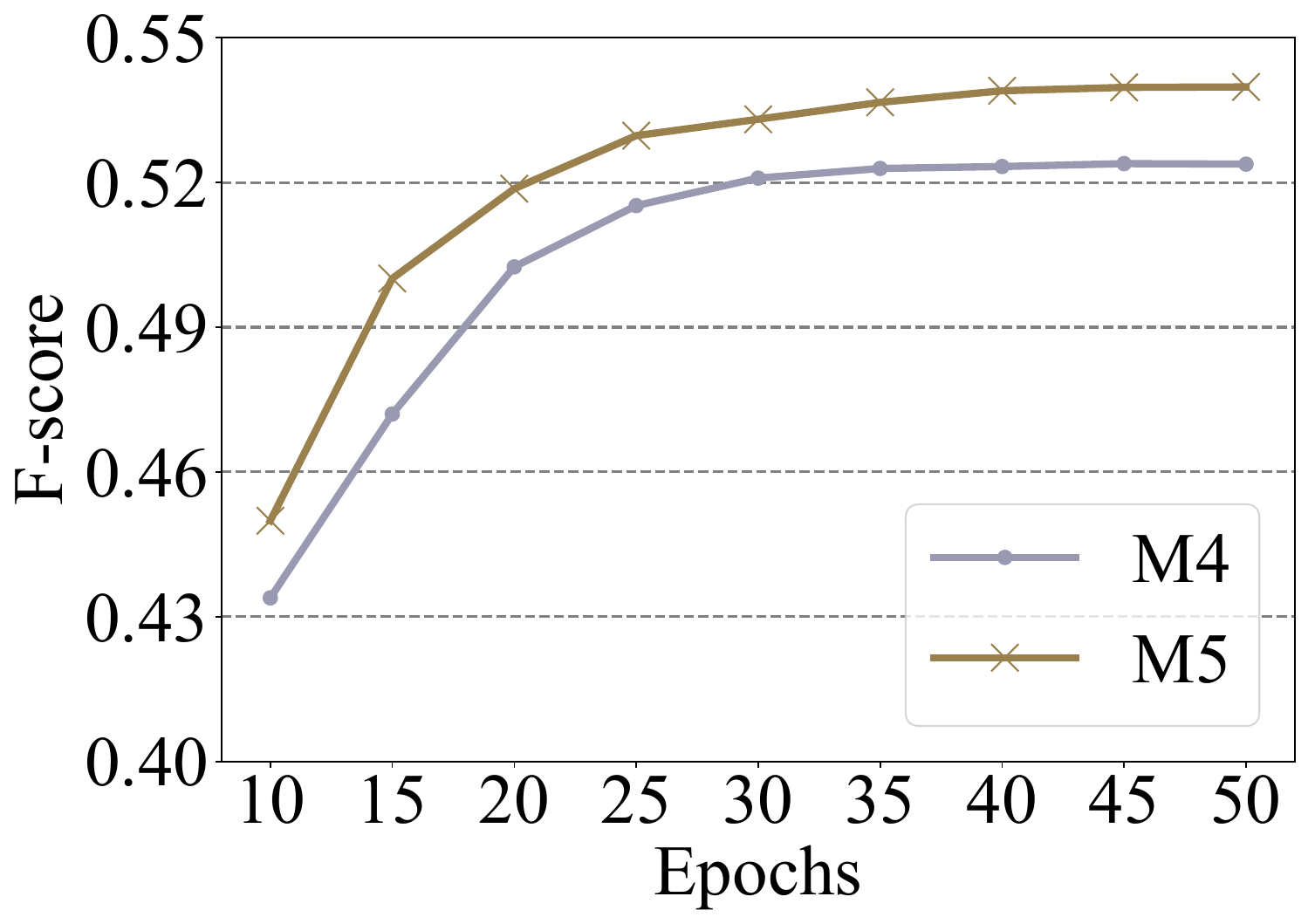}
        \caption{SYSU-MM01}
        \label{f5a}
    \end{subfigure}
    \hfill
    \begin{subfigure}[b]{0.225\textwidth}
        \centering
        \includegraphics[width=\textwidth]{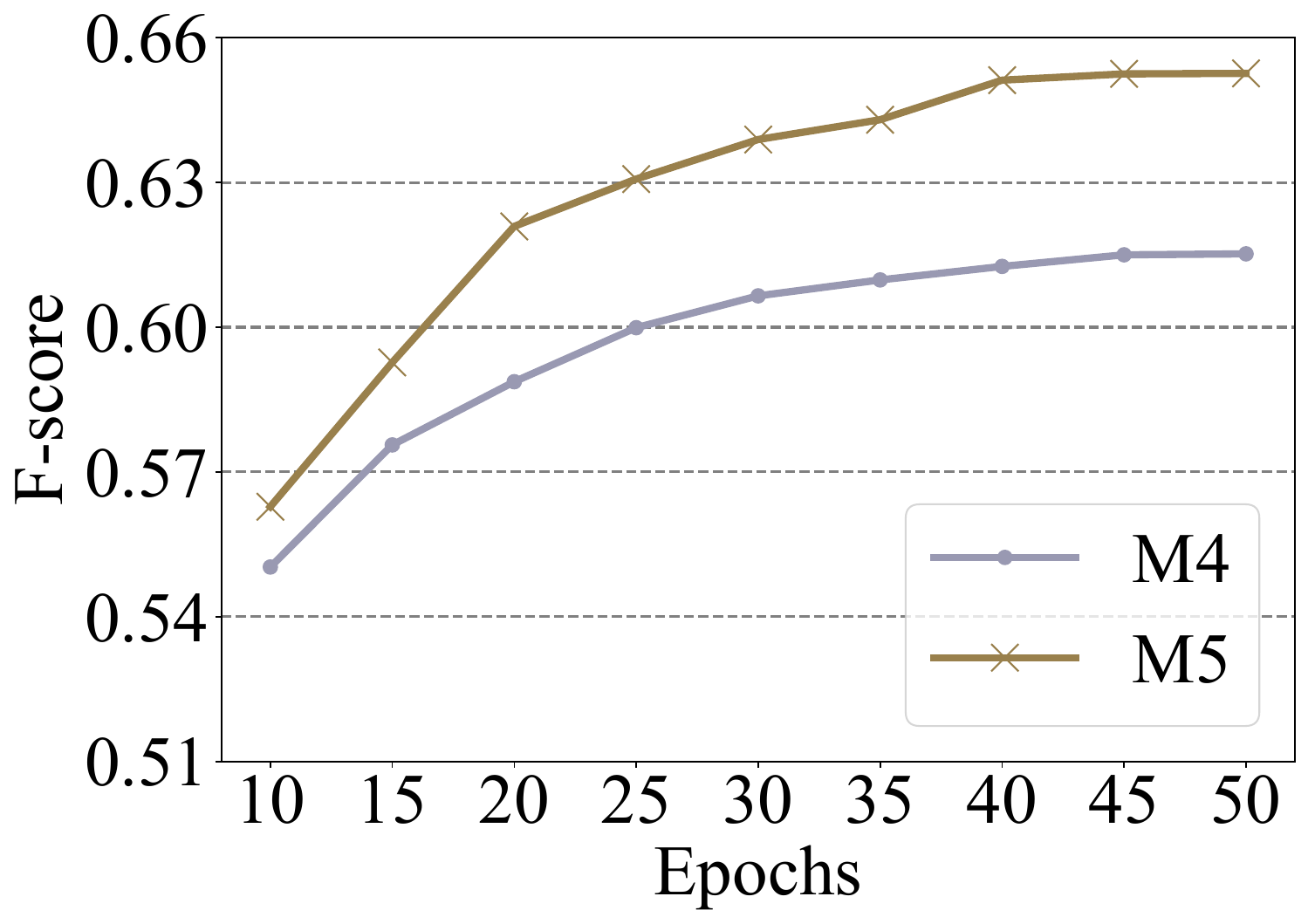}
        \caption{MLR-CUHK03}
        \label{f5b}
    \end{subfigure}
    \caption{F-score for M4 and M5.}
    \label{f5}
\end{figure}

As shown in Table \ref{t5}, M5 achieves significant performance improvements over M4. Additionally, to assess the effectiveness of CHM in obtaining reliable heterogeneous positive pairs, we calculate the F-score \cite{fscore} of the training pseudo-labels for M4 and M5. A higher F-score indicates greater accuracy of the pseudo-labels. As illustrated in Figure \ref{f5}, M5 consistently achieves a higher F-score than M4 throughout the training process. This confirms that CHM enhances model performance by improving the accuracy of heterogeneous positive pairs. 

\begin{figure}[!t]
\centering
\includegraphics[width=0.45\textwidth]{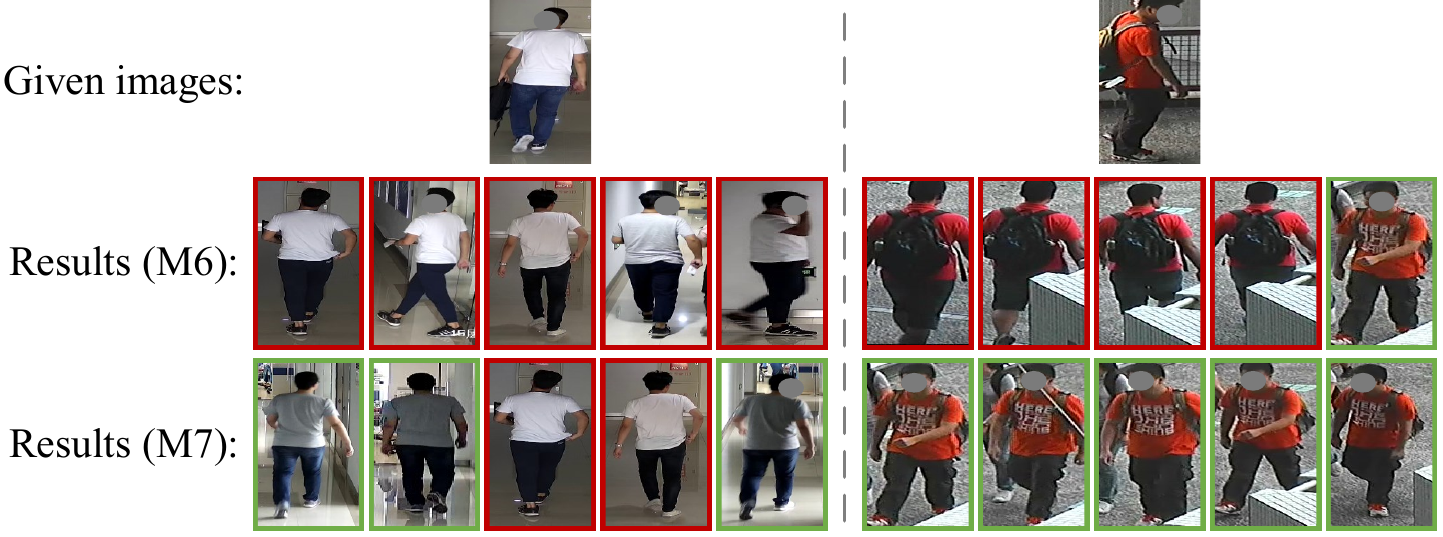}
\caption{Visualization of instance-level heterogeneous positive sets obtained by M6 and M7. True and false positive samples are marked with green and red boxes, respectively.
}
\label{f6}
\end{figure}

\subsubsection{Effectiveness of IHM.}
In Table \ref{t5}, M6 and M7 introduces CNL \cite{sdcl} and IHM, respectively, to obtain instance-level heterogeneous positive sets based on M5, and both achieve performance improvements. This confirms the feasibility of obtaining heterogeneous positive pairs at the instance-level. Furthermore, we find that M7 outperforms M6. To explore this further, we visualize some of the instance-level heterogeneous positive sets obtained by M6 and M7 in Figure \ref{f6}. Compared to M6, M7 effectively excludes some false positive samples that are visually similar. This may be due to the different cognitive focuses of the text encoder and image encoder, allowing IHM to avoid certain misjudgments of the image encoder by relying on consistency. In contrast, although CNL also relies on consistency, specifically between shallow and deep representations of the image encoder, it is inherently dependent on the knowledge within the image encoder, making it difficult to overcome the image encoder's inherent limitations. This validates the effectiveness of IHM. 

\section{Conclusion}
We introduce the UMS-ReID as a new task and propose the ITKM framework to tackle the challenges associated with UMS-ReID. Ablation experiments validate the effectiveness of various components in ITKM, namely the scenario embeddings and multi-scenario separation loss effectively encourage the model to adapt to multiple scenarios, while CHM and IHM obtain reliable heterogeneous positive pairs. Additionally, DRU effectively enhances model performance by maintaining consistency between text and image supervision signals. 
Experimental results validate the superiority and generalizability of ITKM, demonstrating that it not only competes effectively with advanced scenario-specific methods within each scenario, but also leverages knowledge from multiple scenarios to further enhance performance.

\section*{Acknowledgments}
This work is supported by National Key Research and Development Program of China (Grant no. 2023YFC3305003) and the National Natural Science Foundation of China (NSFC, Grant no. 62231013).

\bibliography{ref}

\clearpage
\appendix
\setcounter{section}{0}
\renewcommand{\thesection}{S\arabic{section}}

\setcounter{figure}{0}
\renewcommand{\thefigure}{S\arabic{figure}}

\setcounter{table}{0}
\renewcommand{\thetable}{S\arabic{table}}

\begin{center}
\Huge{\textbf{Supplementary Materials}}
\end{center}

\renewcommand{\algorithmicrequire}{\textbf{Input:}} 
\renewcommand{\algorithmicensure}{\textbf{Output:}}
\begin{algorithm} 
\caption{Image-Text Knowledge Modeling (ITKM)}
\label{alg1} 
\begin{algorithmic}[1]
  
\renewcommand\algorithmiccomment[1]{%
  \hfill\#\ \nb{\footnotesize{#1}}%
}
\REQUIRE Unlabeled training sets $\{ {X^s}\} _{s = 1}^S$,  pre-trained CLIP image and text encoders $E_I^{\textrm{init}}$ and $E_T^{\textrm{init}}$, and $epochs$.
\STATE $E_I \leftarrow E_I^{\textrm{init}}$ with dual front-ends
\FOR {$i = {\rm{ }}1$ to $epochs$}
\STATE Extract representations from $\{ {X^s}\} _{s = 1}^S$ using $E_I$
\STATE Cluster within homogeneous image groups in each scenario to generate homogeneous pseudo-labels
\STATE Optimize $E_I$ based on homogeneous contrastive loss
\ENDFOR
\FOR {$i = {\rm{ }}1$ to $epochs$} 
\STATE Optimize learnable text embeddings ${[X]_1}{\rm{}}[{X_2}]{\rm{}} \dots {\rm{}}[{X_M}]$ to associate each homogeneous pseudo-label with the sentence  ``A photo of a ${[X_1]}{\rm{ }}[{X_2}]{\rm{ }}...{\rm{ }}[{X_M}]$ person'' via a pre-trained CLIP model.
\ENDFOR
\STATE Extract cluster-level text representations using $E_T^{\textrm{init}}$
\STATE Use cluster-level text representations to initialize instance-level text representations
\FOR {$i = {\rm{ }}1$ to $epochs$}
\STATE Extract representations from $\{ {X^s}\} _{s = 1}^S$ using $E_I$
\STATE Clustering within homogeneous image groups in each scenario to generate homogeneous pseudo-labels
\STATE Using DRU, obtain online cluster-level text representations and online instance-level text representations
\STATE Obtain heterogeneous positive clusters using CHM
\STATE Obtain heterogeneous positive sets using IHM
\STATE Optimize $E_I$ to minimize overall loss  ${L_{s3}}$
\ENDFOR
\RETURN Trained $E_I$
\end{algorithmic} 
\end{algorithm}

\section{Homogeneous Group Division}
We divide the heterogeneous images within each scenario into two groups of homogeneous images, labeled as $a$ and $b$. For instance, in UVI-ReID, modality labels are readily accessible, allowing us to easily obtain a group of visible images and a group of infrared images. Similarly, in UCR-ReID, resolution labels are also easy to obtain, allowing us to easily group high-resolution and low-resolution images. For UCC-ReID, since the image encoder is highly sensitive to color and texture information in the initial stages~\cite{cicl}, images of the same person in different outfits often exhibit significant representation differences. Therefore, we extract the representations of all images using the image encoder and perform $K$-means clustering (with $K{\rm{ = }}2$) on all representations to divide the images into two groups.

\section{Dataset Details}
SYSU-MM01 \cite{sysumm01} is a visible-infrared dataset captured by four visible cameras and two infrared cameras. Its training set includes 22,258 visible images and 11,909 infrared images from 395 identities, while the query and gallery sets contain infrared and visible images from 96 identities, respectively. Following existing methods \cite{pgm,sdcl}, we set up two test settings: All Search and Indoor Search. In the All Search setting, the gallery set is captured by four visible cameras, whereas in the Indoor Search setting, the gallery set is captured by only two indoor visible cameras.
\par LTCC \cite{ltcc} is a clothing change dataset, comprising 17,119 images from 152 identities, captured by 12 cameras, with each identity being captured by at least two cameras. The training set, query set, and gallery set contain 9576, 493, and 7050 images, respectively. Following existing methods \cite{sicl,cicl}, we use two test settings: General and Clothing Change settings. The General setting includes both clothing change and consistent clothing images in the gallery set, while the Clothing Change setting includes only clothing change images in the gallery set.
\par MLR-CUHK03 is a cross-resolution dataset based on CUHK03 \cite{cuhk03}, containing images from 1,467 identities captured by two cameras. Following existing works \cite{intact,drfm}, we use 1,367 identities for training and the remaining 100 identities for testing. We downsample the images captured by one camera using a random downsampling rate $\gamma  \in \left\{ {2,3,4} \right\}$ to produce low-resolution (LR) images, while the images captured by the other camera are used as high-resolution (HR) images. The query set and gallery set contain only LR images and HR images, respectively.

\section{Implementation Details}
We use pre-trained image and text encoders from CLIP~\cite{clip}, and design the image encoder $E_I^{\textrm{init}}$ with a frontend consisting of two branches to accommodate all scenarios. 
Data augmentation for the training images includes random flipping and random erasing. 
In each scenario, we set the batch size to $B {\rm{ = }} 64$. 
DBSCAN is employed for clustering within homogeneous groups, with the distance threshold and minimum sample size set to 0.6 and 4, respectively. For the learnable text embeddings ${[X]_1}{\rm{ }}[{X_2}]{\rm{ }}...{\rm{ }}[{X_M}]$, we set $M{\rm{ = }}4$. For CHM, we set $\beta{\rm{ = }}0.5$. For IHM, we set $k {\rm{ = }} 200$. For DRU, we set $\eta {\rm{ = }} 80\%$ and $\alpha  {\rm{ = }} 0.8$. For the temperature hyperparameters, we set ${\tau} {\rm{ = }} 0.05$. The hyperparameter values ${\lambda _{mss}}{\rm{ = }}2.0$  and ${\lambda _{tgc}}{\rm{ = }}1.0$ are used for the experiments. 
Sensitivity analysis of ITKM with respect to ${\lambda _{mss}}$ and ${\lambda _{tgc}}$ for the SYSU-MM01 and LTCC datasets  is provided in Section~\ref{sec:HypParamSensAnalysis} of the Supplementary Materials. 
Each training stage lasts for 50 epochs. We employ the Adam optimizer with an initial learning rate of 0.00035, implementing a warm-up phase for the first 10 epochs. In the test phase, only the final trained image encoder $E_I$ is used for the inference. 
Model training and evaluation are performed on an Ubuntu-based system configured with four NVIDIA Tesla V100 GPUs. The source code can be found on https://github.com/zqpang/ITKM.

\section{Further Comparison}
In theory, existing unsupervised scenario-specific methods, when trained separately on three distinct scenarios and tested by selecting the corresponding weights for each scenario, also have the potential to achieve results similar to those of the UMS-ReID method. To validate this idea, we trained separately the existing advanced method SDCL \cite{sdcl} on the three scenarios and tested the resulting three models in their respective scenarios. We refer to these three models as SDCL(S). As shown in Table~\ref{t4}, the performance of SDCL(S) not only lags behind the performance of ITKM(M), but the number of parameters ($10^9$) required during the testing phase is three times that of ITKM(M). This is because ITKM(M) can integrate knowledge from all three scenarios into a single model. These results not only demonstrate the superiority of the proposed method, but also confirm the research value of UMS-ReID.

\begin{table}
  \centering 
{\fontsize{9}{10}\selectfont
\setlength{\tabcolsep}{3pt}
\begin{tabular}{ccccccc}
\hline
        & All  & Indoor & Change & General & MLR  & Parameters↓ \\ \hline
SDCL(S) & 64.5 & 71.4   & 20.5   & 61.6    & 58.5 & 319.8       \\
SDCL(M) & 63.0 & 69.2   & 18.8   & 55.2    & 35.3 & 106.6       \\
ITKM(M) & 64.9 & 72.3   & 27.3   & 63.3    & 63.6 & 106.6       \\ \hline
\end{tabular}}
\caption{Comparison of ITKM and SDCL. ``All'', ``Indoor'', ``Change'', ``General'', and ``MLR'' represent the Rank-1 accuracy for SYSU-MM01 (All Search), SYSU-MM01 (Indoor Search), LTCC (Clothing Change), LTCC (General), and MLR-CUHK03, respectively\label{t4}}
\end{table}

\section{Sensitivity Analysis} \label{sec:HypParamSensAnalysis}
\begin{figure}[h]
    \centering
    \begin{subfigure}[b]{0.215\textwidth}
        \centering
        \includegraphics[width=\textwidth]{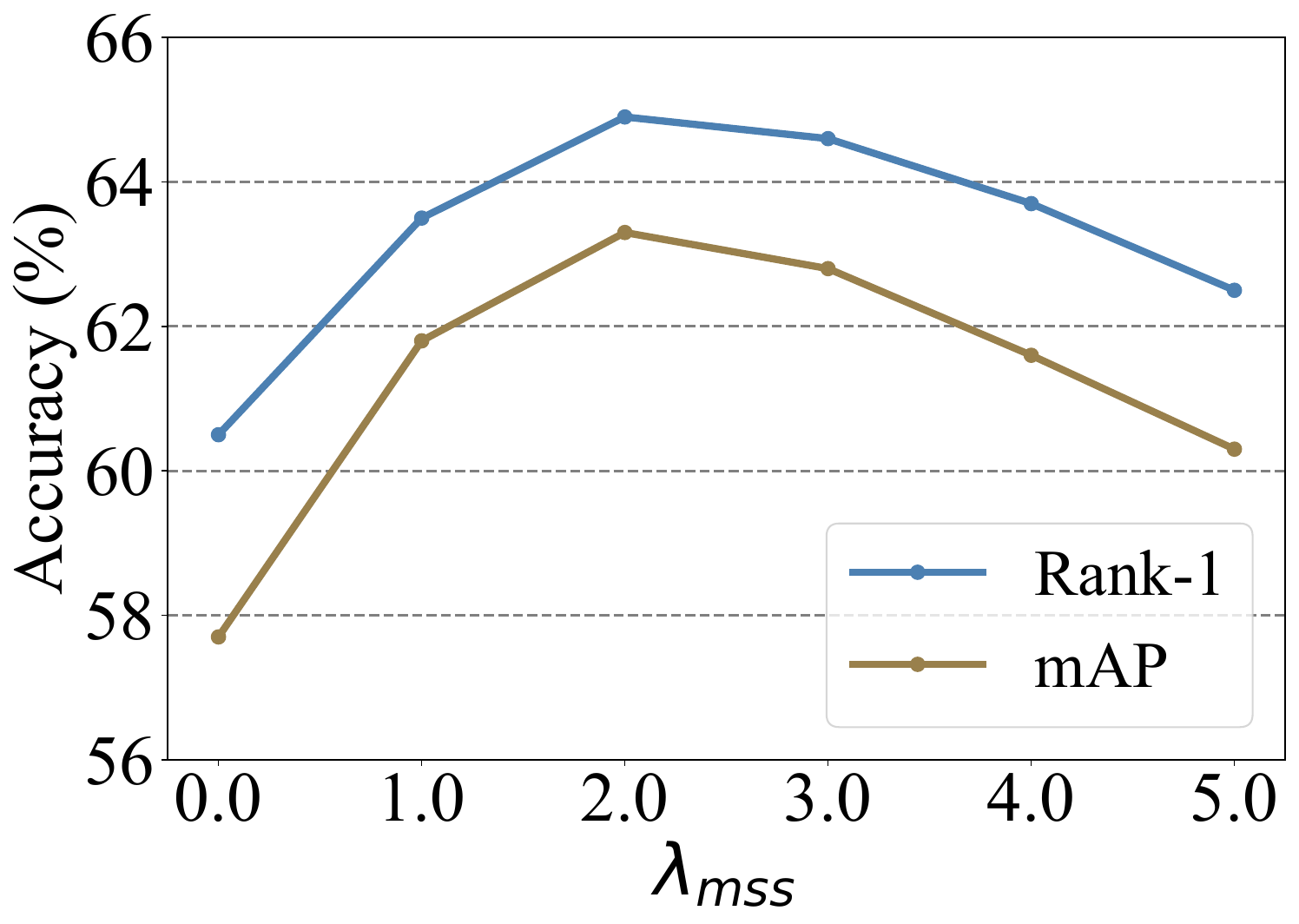}
        \caption{SYSU-MM01 (All Search)}
        \label{s1a}
    \end{subfigure}
    \hfill
    \begin{subfigure}[b]{0.215\textwidth}
        \centering
        \includegraphics[width=\textwidth]{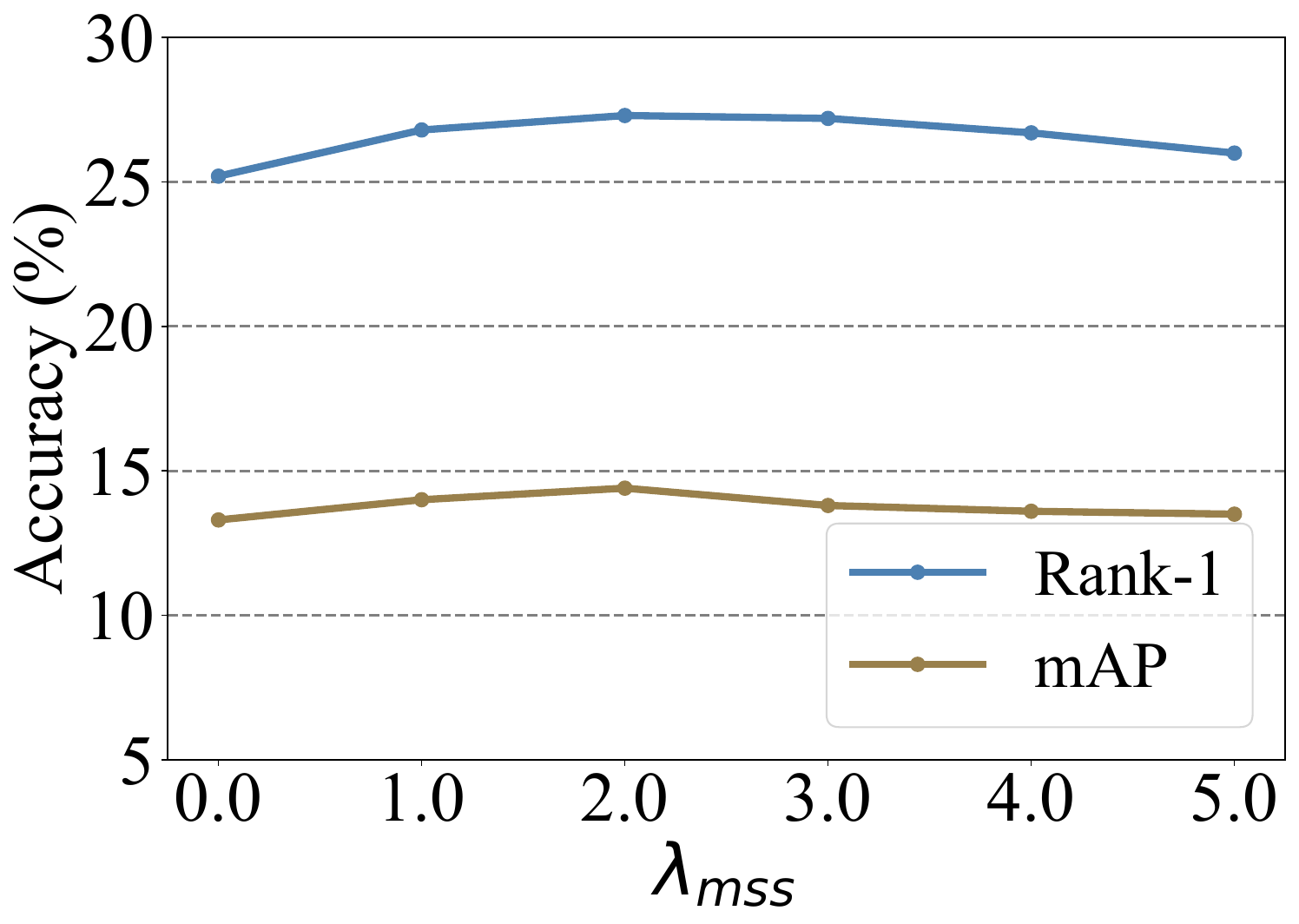}
        \caption{LTCC (Clothing Change)}
        \label{s1b}
    \end{subfigure}
    \caption{Impact of hyperparameter ${\lambda _{mss}}$ on performance.}
    \label{s1}
\end{figure}

Figure~\ref {s1} shows how the model performance (Rank-1 and mAP) varies as ${\lambda _{mss}}$ is varied from 0.0 to 5.0. For ${\lambda _{mss}}{\rm{ = }}0.0$,  ${L_{mss}}$ does not contribute to the overall loss. We observe that when ${\lambda _{mss}}{\rm{ = }}0.0$, the model achieves the worst performance, which preliminarily verifies the effectiveness of ${L_{mss}}$. We also find that the model obtains optimal performance on both datasets when ${\lambda _{mss}}{\rm{ = }}2.0$, thus validating the generalizability of ${\lambda _{mss}}$.

\par
\begin{figure}[h]
    \centering
    \begin{subfigure}[b]{0.215\textwidth}
        \centering
        \includegraphics[width=\textwidth]{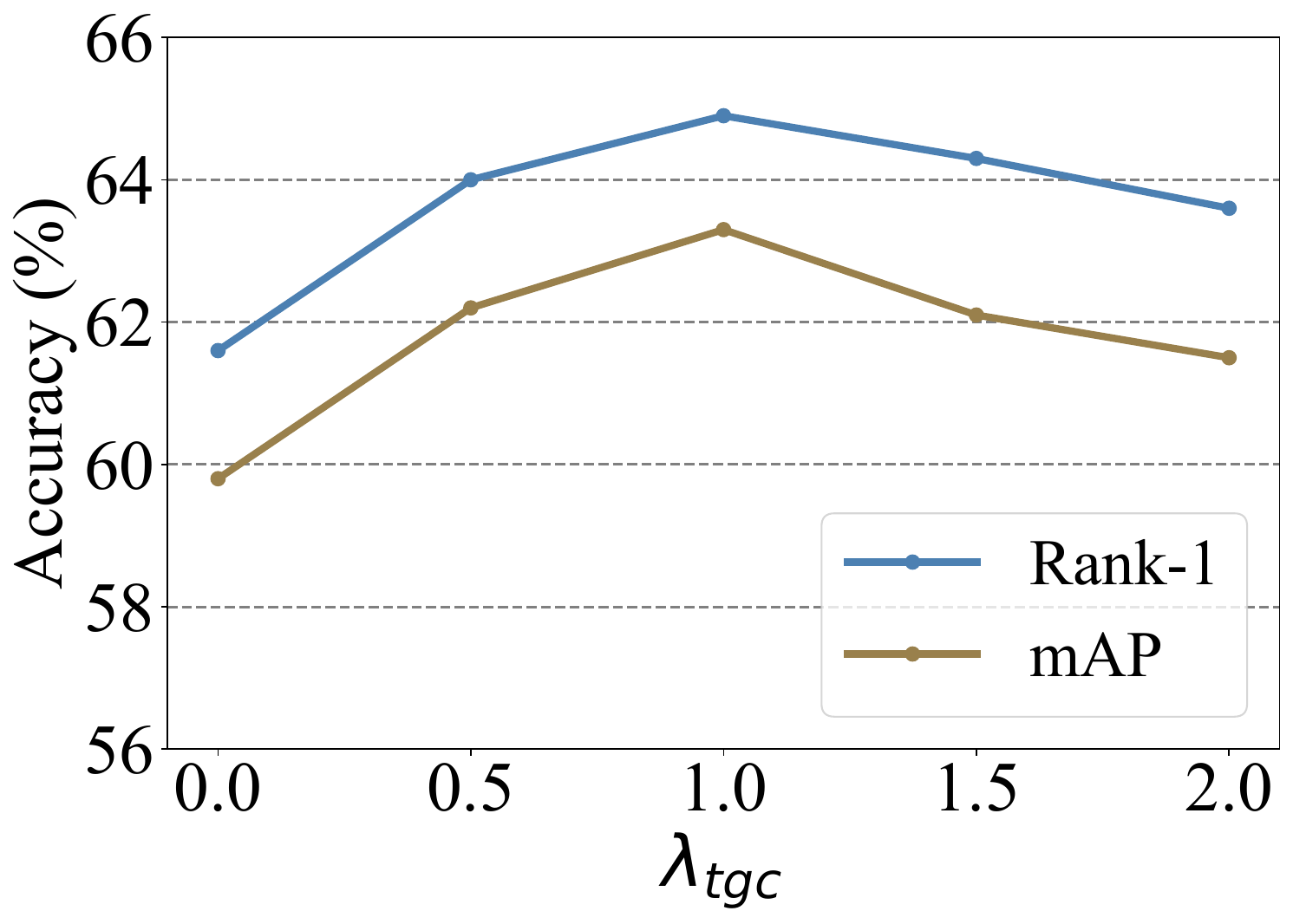}
        \caption{SYSU-MM01 (All Search)}
        \label{s2a}
    \end{subfigure}
    \hfill
    \begin{subfigure}[b]{0.215\textwidth}
        \centering
        \includegraphics[width=\textwidth]{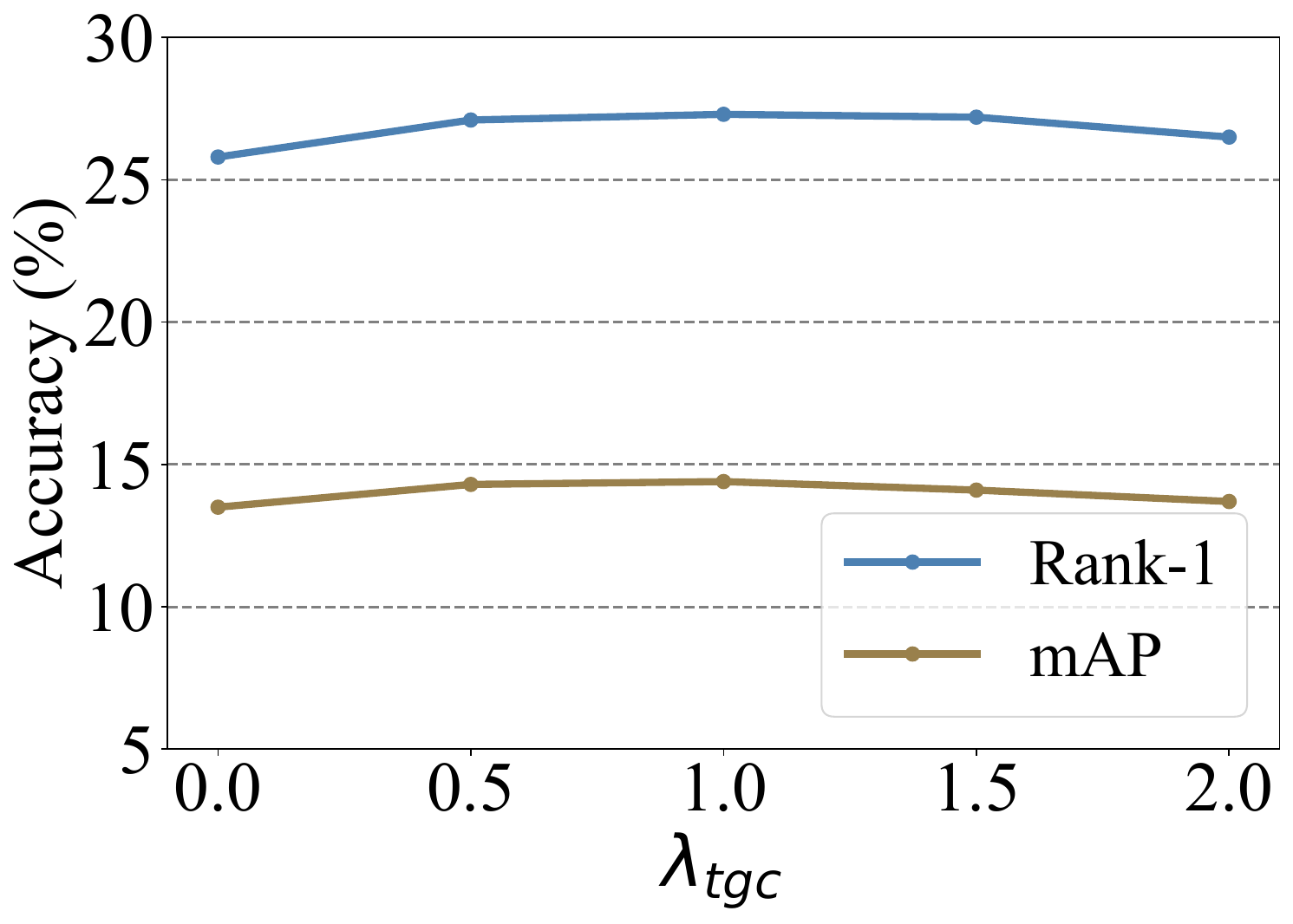}
        \caption{LTCC (Clothing Change)}
        \label{s2b}
    \end{subfigure}
    \caption{Impact of hyperparameter ${\lambda _{tgc}}$ on performance.}
    \label{s2}
\end{figure}

 In Figure~\ref {s2}, we investigate the optimal setting for ${\lambda _{tgc}}$ on two datasets. The model achieves the best performance when ${\lambda _{tgc}}{\rm{ = }}1.0$. When ${\lambda _{tgc}}$ is set to either a larger or smaller value, the performance declines. This is because a small ${\lambda _{tgc}}$ reduces the contribution of ${L_{tgc}}$, while a large ${\lambda _{tgc}}$ diminishes the effectiveness of CHM and IHM.

\par
\begin{figure}[h]
    \centering
    \begin{subfigure}[b]{0.215\textwidth}
        \centering
        \includegraphics[width=\textwidth]{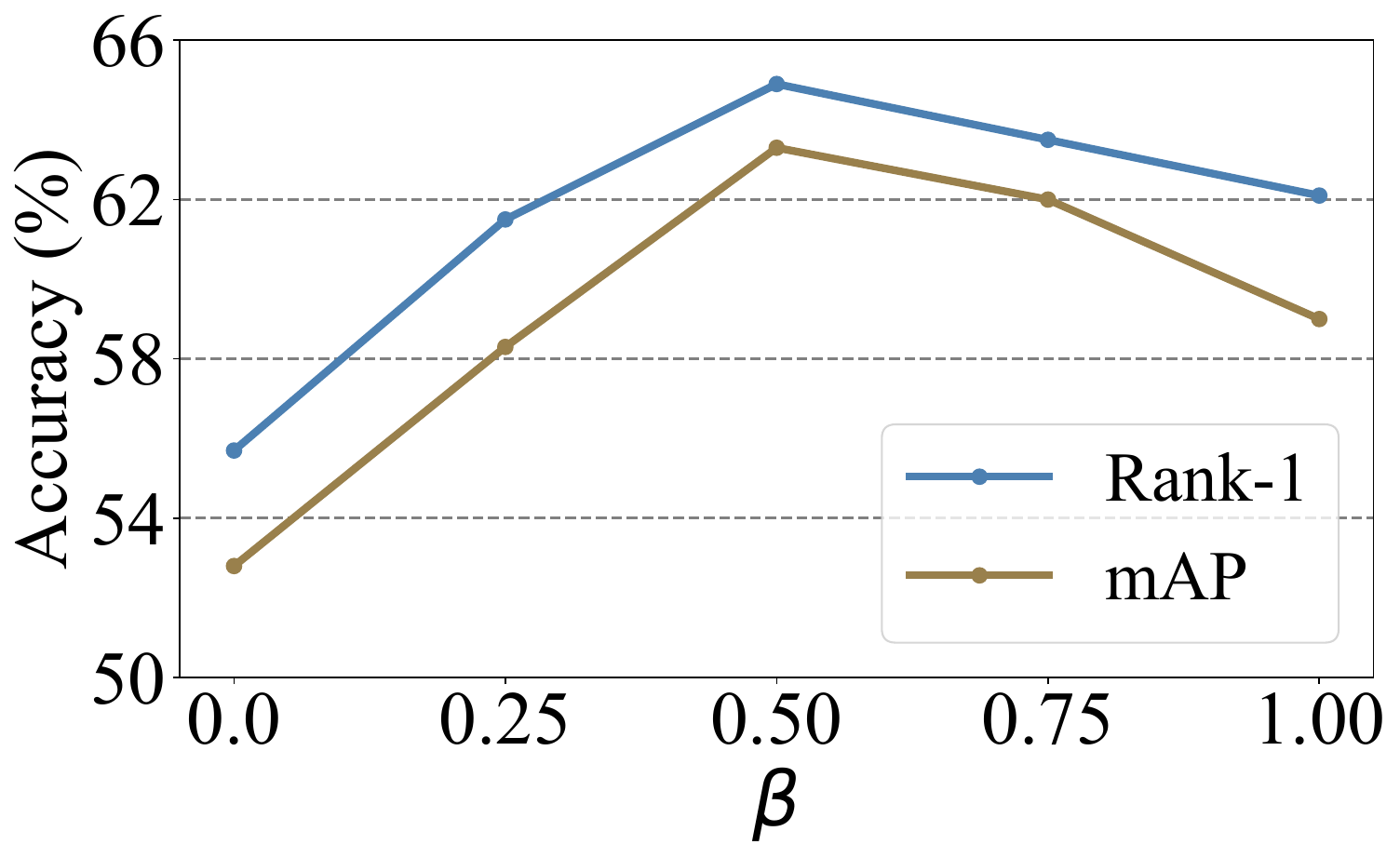}
        \caption{SYSU-MM01 (All Search)}
        \label{s3a}
    \end{subfigure}
    \hfill
    \begin{subfigure}[b]{0.215\textwidth}
        \centering
        \includegraphics[width=\textwidth]{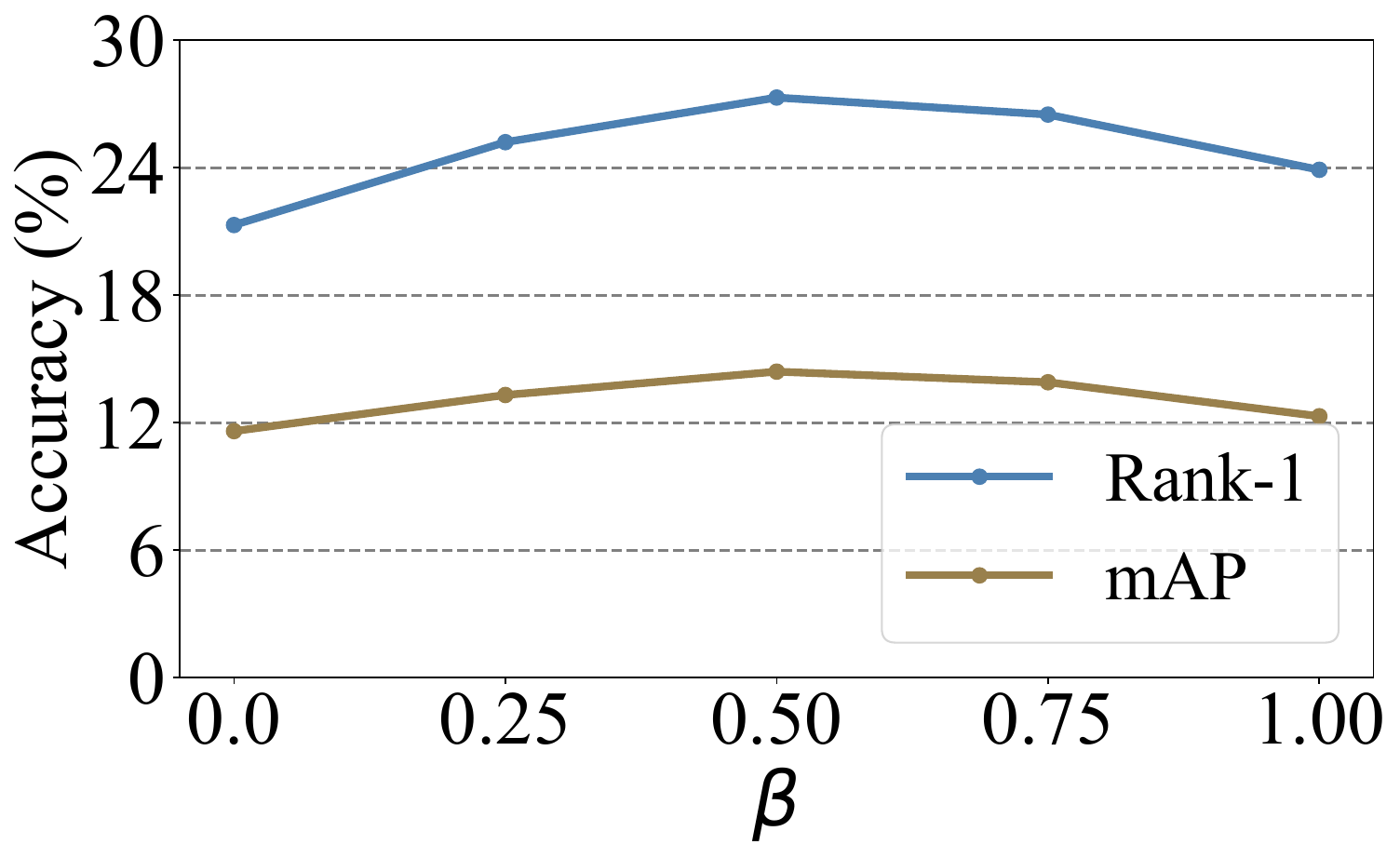}
        \caption{LTCC (Clothing Change)}
        \label{s3b}
    \end{subfigure}
    \caption{Impact of hyperparameter $\beta $ on performance.}
    \label{s3}
\end{figure}

 We retain inconsistent results with a probability of $\beta$ because there are a significant number of such results (approximately 50\%), and not all of them are necessarily incorrect. Discarding all of them would negatively affect the data diversity. As shown in Figure~\ref{s3}, a low $\beta$ leads to poorer performance, while a very high $\beta$ also causes performance degradation. This is because a high $\beta$ retains a large number of unreliable matches, which misguides the optimization.

\section{Further Analysis}

\begin{figure}[h]
    \centering
    \begin{subfigure}[b]{0.225\textwidth}
        \centering
        \includegraphics[width=\textwidth]{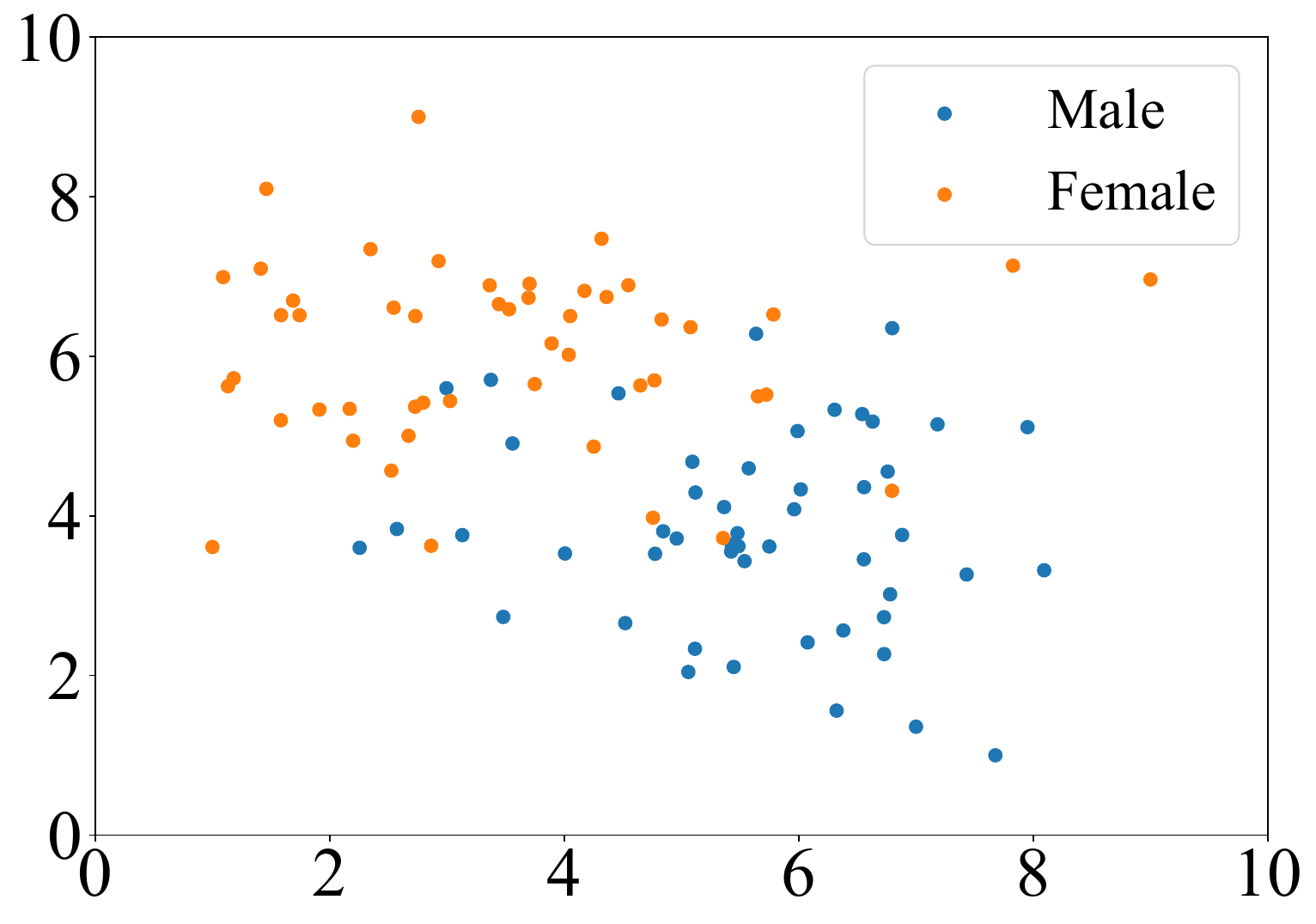}
        \caption{SYSU-MM01}
        \label{s4a}
    \end{subfigure}
    \hfill
    \begin{subfigure}[b]{0.225\textwidth}
        \centering
        \includegraphics[width=\textwidth]{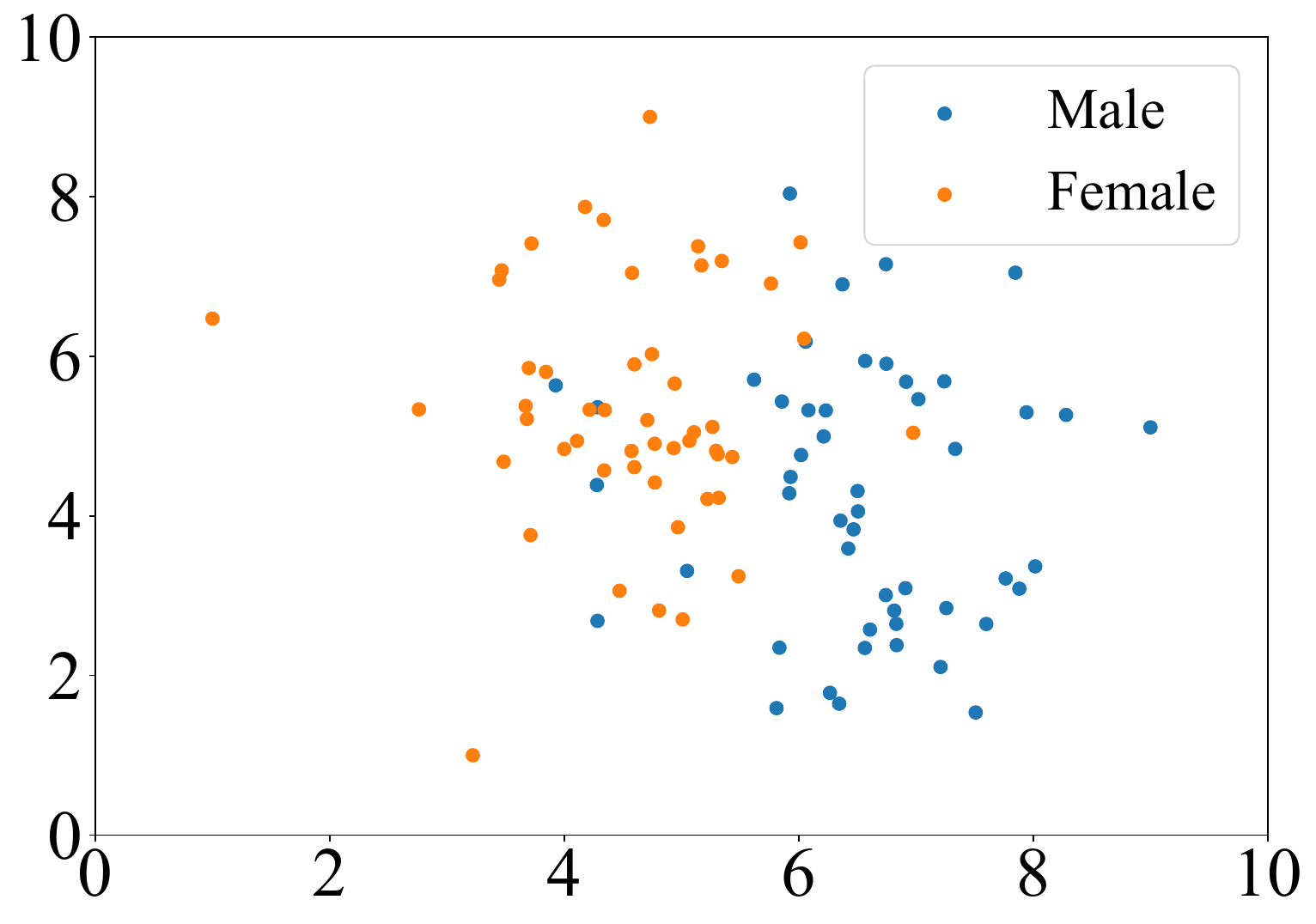}
        \caption{LTCC}
        \label{s4b}
    \end{subfigure}
    \caption{Two-dimensional t-SNE \cite{tsne} visualizations of the text representations.}
    \label{s4}
\end{figure}

The motivation for using vision language models in ITKM is to exploit the powerful representation power of vision language models, specifically, to leverage the ability of language to represent identity related information at a higher-level of semantics that are likely to be invariant across heterogeneous groups and across scenarios. We therefore examined the learnable text embeddings to determine whether they indeed capture high-level identity-related semantic information, and found that to indeed be the case. As an example, we visualize the text representations on SYSU-MM01 and LTCC datasets, respectively, as shown in Figure~\ref{s4}. The distributions of text representations in both scenarios appear to be correlated with gender. This verifies the effectiveness of the text representations, indicating that they not only guide the image encoder to adapt to different scenarios, but also encode identity-relevant semantic information, which has the potential to facilitate the identity-related learning of the image encoder. 

Further analysis to determine how the learned text representations capture other high level semantics is clearly of interest, among other things for endowing explainability to the models, but beyond the scope of this paper.


\end{document}